\documentclass[11pt]{article}

\usepackage[preprint]{acl}

\usepackage{times}
\usepackage{latexsym}

\usepackage[T1]{fontenc}

\usepackage[utf8]{inputenc}

\usepackage{microtype}

\usepackage{inconsolata}

\usepackage{graphicx}

\usepackage{threeparttable}
\usepackage[table]{xcolor}
\usepackage[most]{tcolorbox} 
\usepackage{xcolor}          
\usepackage{booktabs}
\definecolor{uciblue}{HTML}{0064A4}
\usepackage{multirow}
\usepackage{amsmath}
\usepackage{amssymb} 
\usepackage{titletoc}
\usepackage{bbm}
\newcommand{\codebf}[1]{\textbf{\texttt{#1}}}

\newtcblisting[auto counter]{promptbox}[2][]{%
  enhanced,
  breakable,
  colback=uciblue!5,
  colframe=black,
  colbacktitle=uciblue!15,
  coltitle=black,
  fonttitle=\bfseries,
  title={Prompt~\thetcbcounter: #2},
  label={#1},
  boxrule=0.6pt,
  arc=6pt,
  outer arc=6pt,
  left=6pt,
  right=6pt,
  top=6pt,
  bottom=6pt,
  listing only,
  listing options={
    basicstyle=\ttfamily\footnotesize\raggedright,
    breaklines=true,
    columns=fullflexible,
    keepspaces=true,
    showstringspaces=false,
    escapeinside={(*@}{@*)}
  }
}

\newcommand{\dname}{DarkNav-Bench}
\newcommand{\mname}{ADAPT}

\title{\mname: \underline{A}udio-\underline{D}epth \underline{A}ction \underline{P}lanning with \underline{T}ransition-Aware Rewards for Lights-Out Industrial Navigation}

\author{
 \textbf{Wenjun Huang\textsuperscript{1}},
 \textbf{Shenghao Fu\textsuperscript{2}},
 \textbf{Yian Jin\textsuperscript{3}},
 \textbf{Yang Ni\textsuperscript{4}},
 \textbf{Yifan Li\textsuperscript{5}},\\
 \textbf{Lichi Li\textsuperscript{6}},
 \textbf{Xingjian Diao\textsuperscript{7}},
 \textbf{Pengfei Zhang\textsuperscript{1}},
 \textbf{Yutong Song\textsuperscript{1}},
 \textbf{Hanning Chen\textsuperscript{1}},\\
 \textbf{Jiarui Cui\textsuperscript{8}},
 \textbf{Tianyu Yang\textsuperscript{9}},
 \textbf{Shilong Li\textsuperscript{1}},
 \textbf{Yu Kong\textsuperscript{5}},
 \textbf{Mohsen Imani\textsuperscript{1}}
\\
\\
 \textsuperscript{1}University of California, Irvine,
 \textsuperscript{2}University of Pennsylvania,
 \textsuperscript{3}Carnegie Mellon University,\\
 \textsuperscript{4}Purdue University Northwest,
 \textsuperscript{5}Michigan State University,
\textsuperscript{6}Independent Researcher,\\
\textsuperscript{7}Dartmouth College,
\textsuperscript{8}University of California, Los Angeles,
\textsuperscript{9}University of Notre Dame,
}

\begin{document}
\maketitle
\begin{abstract}
Vision-language-action models for embodied navigation typically rely on RGB observations, limiting their applicability in \emph{lights-out} industrial environments where visible illumination may be unavailable.
We study illumination-independent navigation using spatial audio for target localization and depth for obstacle-aware geometry.
To support this setting, we introduce \dname, a benchmark with 24 industrial scenes and four progressively challenging acoustic conditions.
We further propose \mname, an audio-depth action-planning framework equipped with spatial-audio perception and optimized for closed-loop navigation through reinforcement learning (RL).
Our key RL contribution, transition-aware reward shaping (TARS), defines a pose-aware shortest-action potential over a position--heading graph. 
Unlike conventional geodesic-progress rewards, TARS accounts for both feasible movements and in-place rotations, providing action-consistent feedback for policy optimization.
Experiments show that \mname{} consistently outperforms training-based navigation baselines and zero-shot multimodal models on both seen and unseen scenes.
On unseen environments, it achieves 90.6\% success under Clean conditions and retains 60.0\% success under the most challenging Localized Noise setting.
These results demonstrate the potential of spatial audio and depth as complementary modalities for reliable RGB-free navigation in lights-out industrial environments.
\end{abstract}

\section{Introduction}
\label{sec:intro}

The rapid adoption of autonomous robots is transforming modern manufacturing, where embodied agents are increasingly expected to operate with minimal human intervention.
Recent advances in large multimodal models (LMMs) have enabled vision-language-action (VLA) models to integrate perception, reasoning, and action generation~\cite{black2024pi_0, kim2024openvla, wang2026qwen, driess2026knowledge}, with promising results for embodied navigation~\cite{hirose2025omnivla, hirose2026asyncvla}.

\begin{figure}[tb]
\centering
\resizebox{\columnwidth}{!}{
\includegraphics{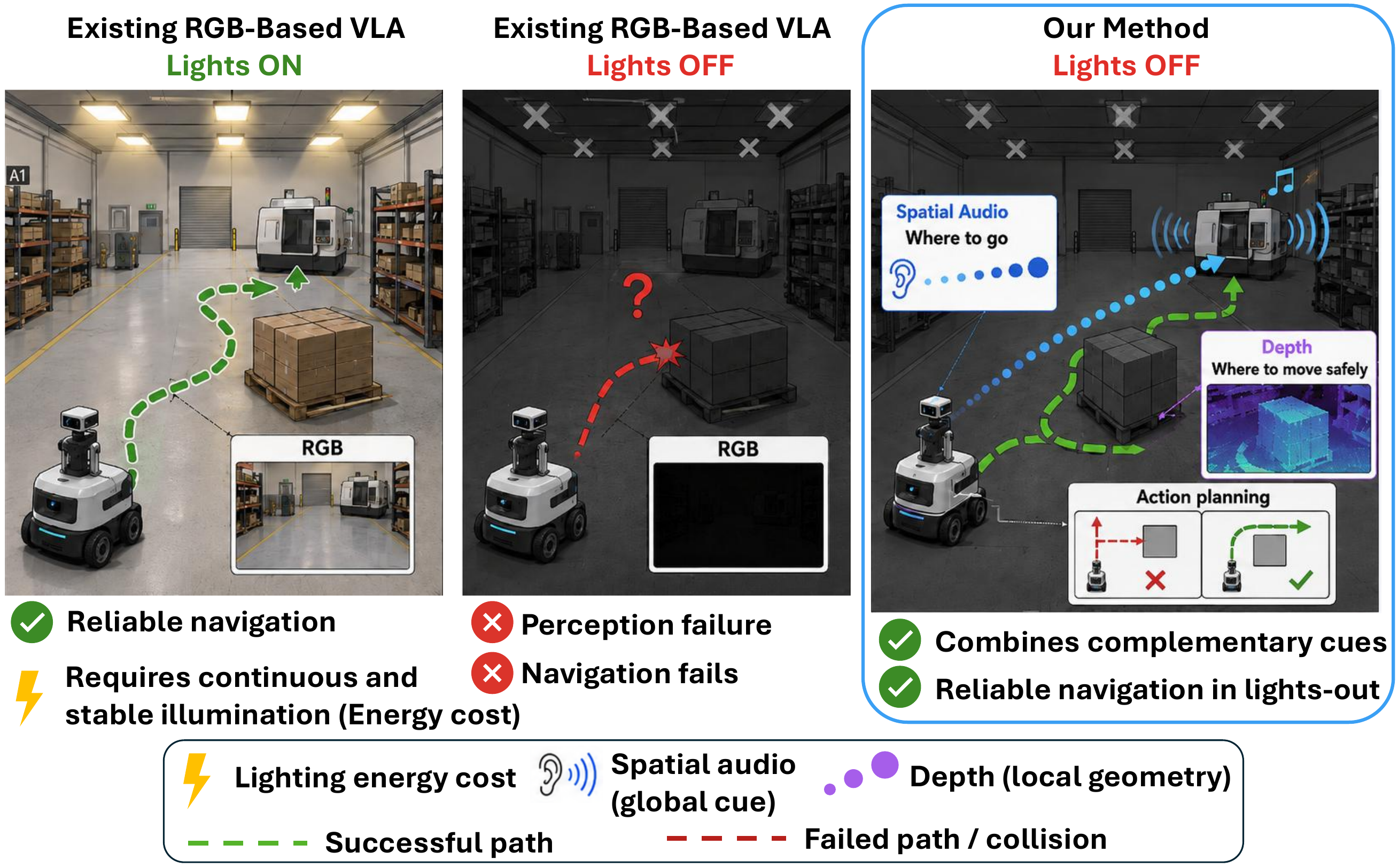}
}
\vspace{-2em}
\caption{
\textbf{Motivation of \mname.} Existing RGB-based VLA agents depend on visible illumination and therefore require continuous lighting in lights-out factories. \mname{} instead combines spatial audio for target direction with depth sensing for local obstacle geometry, enabling reliable and energy-efficient navigation without visible light.
}
\vspace{-2em}
\label{fig:teaser}
\end{figure}

However, most existing approaches rely heavily on RGB observations, implicitly assuming that the environment remains sufficiently and consistently illuminated.
This assumption becomes problematic in increasingly automated industrial environments~\cite{li2026industryeqa}.
As manufacturing moves toward \emph{lights-out operation}~\cite{standardbots2026lightsout, karp2025lightsout}, factories can operate with little or no human presence and therefore no longer require continuous visible illumination.
Maintaining sufficient lighting primarily for robotic perception introduces unnecessary energy consumption, while reducing illumination severely degrades RGB-based navigation systems.
Consequently, existing VLA agents may become unreliable or even unusable in lights-out factories.
This motivates an important question: \emph{\textbf{can embodied agents navigate reliably without relying on visible-light perception?}}

Spatial audio and depth provide complementary cues for this setting.
Multi-channel audio can reveal the direction of an acoustic target regardless of illumination~\cite{holland2002audiogps, wilson2007swan}, but provides limited information about obstacles and traversable space.
Depth sensing, in contrast, captures local geometry and remains effective without visible illumination~\cite{zeng2025depthdark, singh2026nightsight}.
Thus, \emph{\textbf{spatial audio indicates where to go, while depth indicates where the agent can safely move}}.
Despite this natural complementarity, their joint use for VLA-based navigation remains largely unexplored.

To study this problem, we introduce \dname, a Unity~\cite{unity}-based benchmark for \textbf{spatial-audio-and-depth navigation in lights-out industrial environments}.
\dname{} contains diverse factory scenes in which an embodied agent must navigate toward an acoustic target using only spatial audio and egocentric depth.
It further evaluates robustness under \textbf{progressively challenging acoustic conditions}, including geometry-dependent propagation, diffuse noise, and localized machinery noise.
We further propose \mname, an LMM for jointly reasoning over spatial audio and depth.
Because existing audio-capable foundation models are primarily designed for monaural audio, we augment the audio pathway with a lightweight spatial-audio adapter that enables directional reasoning.
The adapted model is then trained to jointly reason over spatial audio and depth for discrete navigation actions.

Finally, we improve navigation with RL.
Existing audio-guided navigation methods commonly use geodesic progress to shape navigation policies~\cite{chen2020soundspaces, zeng2026semantic}. 
Although geodesic distance captures obstacle-aware path length, it depends only on the agent's position and is therefore insensitive to in-place rotations. 
However, turning toward a feasible route and turning away from it may receive identical geodesic feedback despite having different consequences for future actions.
We therefore introduce TARS, which defines a pose-aware potential over a position--heading action graph. 
TARS measures the minimum remaining number of executable navigation actions, explicitly accounting for both forward movements and rotations. TARS thus aligns local reward feedback with the agent's motion primitives and long-horizon navigation progress, rather than geometric path length alone.

Our contributions are threefold:
\begin{itemize}
    \item We introduce \dname, an RGB-free industrial navigation benchmark using spatial audio and depth under progressively challenging acoustic conditions.
    \item We propose \mname, an LMM adapted to spatial-audio-and-depth navigation via a lightweight spatial-audio adapter and multimodal action fine-tuning.
    
    \item We develop TARS, a pose-aware and action-consistent reward potential computed from shortest-action distances on a position--heading graph, enabling RL to distinguish beneficial from counterproductive rotations overlooked by geodesic-progress rewards.
    
\end{itemize}

\section{Task Formulation}
\label{sec:task_formulation}

We formulate lights-out industrial navigation as a partially observable sequential decision-making problem.
An embodied agent navigates toward an active acoustic target while avoiding obstacles, using only spatial audio and egocentric depth observations.
Unlike conventional vision-based navigation, the agent has no access to RGB observations or visible illumination.
At timestep $t$, the underlying simulator state is denoted by $S_t$, which contains the robot pose, target location, and environment geometry.
These variables are used only for simulation and supervision and are \emph{\textbf{not}} exposed to the policy.
Instead, the agent observes
\begin{equation}
    O_t =
    \left(
        \mathbf{X}^{a}_t,\,
        \mathbf{X}^{d}_t
    \right),
    \label{eq:observation}
\end{equation}
where $\mathbf{X}^{a}_t$ is a four-channel first-order ambisonics (FOA) observation and $\mathbf{X}^{d}_t$ is an egocentric depth map.
Spatial audio provides directional cues toward the active target, while depth captures local scene geometry for collision-aware motion.

We use a discrete action space
\begin{equation}
    \mathcal{A}
    =
    \{
        a^{\mathrm{fwd}},
        a^{\mathrm{left}},
        a^{\mathrm{right}},
        a^{\mathrm{stop}}
    \},
    \label{eq:action_space}
\end{equation}
corresponding to moving forward, turning left, turning right, and stopping. 
At timestep $t$, the policy selects an action conditioned on the
information available up to the current step:
\begin{equation}
    \pi_\theta(a_t \mid h_t),
    \label{eq:policy}
\end{equation}
where $h_t$ denotes the policy's available interaction context,
including the current observation $O_t$ and, when used, past actions.

Let $\mathbf{p}_t$ and $\mathbf{p}^{s}$ denote the robot and sound-source positions, respectively.
The target distance is
\begin{equation}
    d_t^{s}
    =
    \left\|
        \mathbf{p}^{s}-\mathbf{p}_t
    \right\|_2,
\end{equation}
and the agent reaches the goal when
\begin{equation}
    d_t^{s} \leq \epsilon,
    \label{eq:euclidian_dist}
\end{equation}
where $\epsilon$ is the success threshold.
An episode terminates when the agent reaches the target and stops, or when the maximum navigation horizon $T_{\max}$ is reached.
The task therefore requires the policy to infer target direction from spatial audio, perceive traversable geometry from depth, and integrate both modalities for sequential collision-aware navigation.
Additional formal definitions are provided in App.~\ref{app:task_formulation}.

\section{Benchmark Construction}
\label{sec:benchmark}

We construct \dname{} by extending IndustryNav~\cite{li2025industrynav} with illumination-independent spatial audio perception.
IndustryNav provides high-fidelity Unity industrial environments with cluttered layouts, dynamic workers and equipment, collision-aware interaction, and egocentric depth sensing.
We retain these geometric and dynamic properties while replacing the original visual navigation setting with an audio-depth navigation task.

\paragraph{Lights-Out Industrial Navigation}
The navigation policy receives only a four-channel FOA observation and an egocentric depth map.
RGB observations may still be rendered for visualization and debugging, but are never provided to the agent.

\paragraph{Acoustic Navigation Target}
Each navigation goal is associated with a stationary acoustic emitter that continuously plays a controlled chirp signal. 
We integrate a modified Steam Audio pipeline~\cite{valve_steam_audio} to spatialize the source relative to an FOA receiver attached to the agent.
As the agent moves and rotates, the received target signal changes according to the relative source--listener configuration, providing a continuously evolving directional cue toward the navigation target.
Depending on the benchmark difficulty, we progressively enable additional acoustic effects and environmental sound sources.

\paragraph{Progressive Acoustic Difficulty}
To systematically evaluate robustness under increasingly realistic acoustic conditions, we construct four progressively challenging settings: \codebf{Clean}, \codebf{Propagation} (\codebf{Prop}), \codebf{Diffuse Noise} (\codebf{DN}), and \codebf{Localized Noise} (\codebf{LN}).
The settings are cumulative, such that each level retains all acoustic factors introduced at the preceding levels.
\codebf{Clean} represents an idealized acoustic environment in which the navigation target is the only sound source and no additional geometry-dependent acoustic effects are introduced.
\codebf{Prop} extends \codebf{Clean} by enabling geometry- and material-dependent sound propagation, including reflection, occlusion, and transmission, causing the target signal to interact with the surrounding industrial scene. 
\codebf{DN} further introduces a spatially diffuse noise field throughout the environment to model persistent ambient acoustic interference.
Finally, \codebf{LN} adds localized, object-associated sound sources on top of the diffuse background, such as operating machinery and sounds generated by moving robots or equipment.
The four settings therefore form a controlled progression
\begin{equation}
    \codebf{Clean}
    \subset
    \codebf{Prop}
    \subset
    \codebf{DN}
    \subset
    \codebf{LN},
\end{equation}
ranging from ideal acoustic localization to navigation under complex and spatially structured industrial interference.

\paragraph{Diffuse Background Noise}
For the \codebf{DN} and \codebf{LN} settings, we construct a spatially diffuse background field
$\mathbf{X}^{\mathrm{bg}}_t$
using real machine recordings from MIMII~\cite{purohit2019mimii}.
We control the strength of the background interference using the signal-to-noise ratio (SNR), defined as the ratio between the power of the target sound and that of the diffuse background noise.
A lower SNR corresponds to stronger background interference.

\paragraph{Localized Noise}
The most challenging \codebf{LN} setting additionally introduces localized acoustic interference $\textbf{X}_t^{\mathrm{item}}$ associated with individual scene entities.
Each item sound has a specific source location in the environment and is spatialized according to its relative position to the agent.
These sources create spatially structured distractors that may overlap with or compete against the target signal, requiring the agent to distinguish the navigation target from other localized acoustic events.
The resulting spatial-audio observation can be expressed as
\begin{equation}
    \mathbf{X}^{a}_t
    =
    \mathbf{X}^{\mathrm{tar}}_t
    +
    \mathbf{X}^{\mathrm{bg}}_t
    +
    \mathbf{X}^{\mathrm{item}}_t,
    \label{eq:audio_mix}
\end{equation}
where the background and item components are included only in their corresponding difficulty levels.

\begin{figure*}[t]
\centering
\resizebox{.95\textwidth}{!}{
\includegraphics{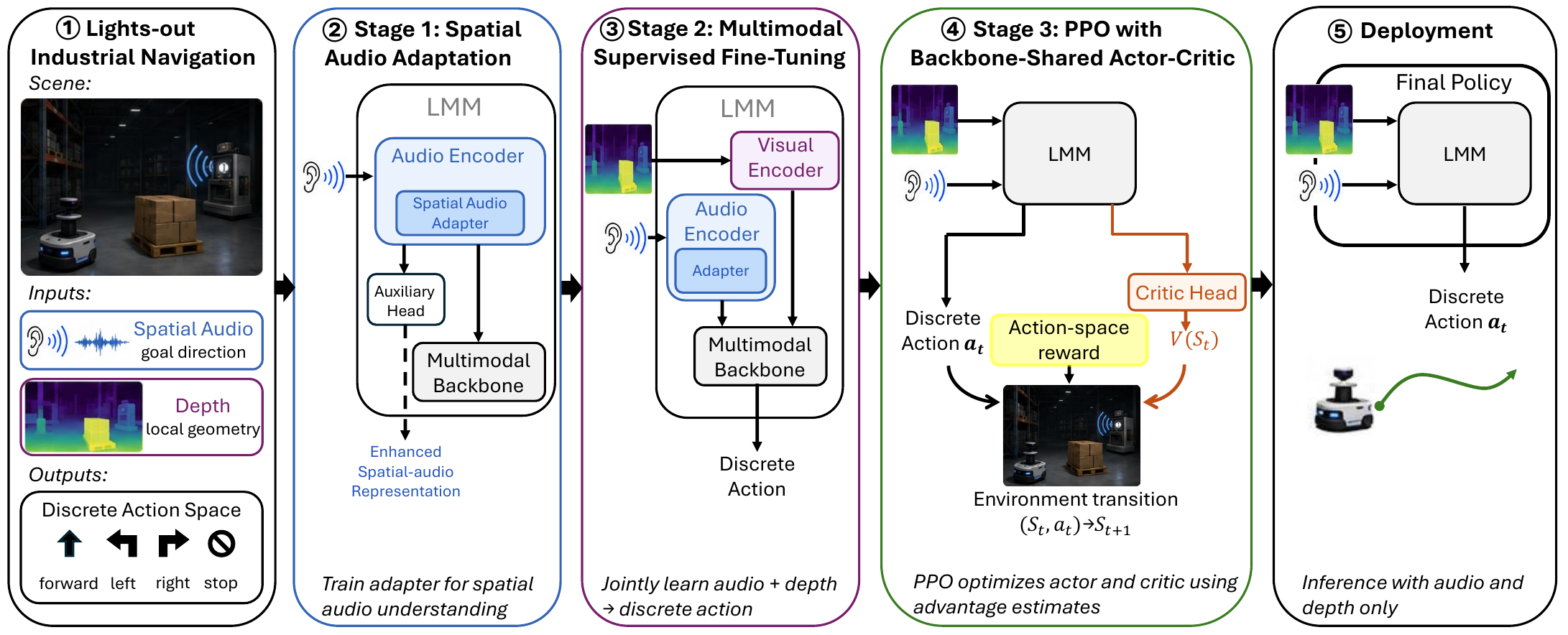}
}
\vspace{-1em}
\caption{
\textbf{Overview of \mname}. The model is trained in three stages: spatial-audio adaptation, multimodal SFT for audio-depth action prediction, and PPO-based closed-loop optimization with TARS. At inference time, the final policy directly maps spatial audio and depth observations to discrete navigation actions.
}
\vspace{-1.5em}
\label{fig:overview}
\end{figure*}

\paragraph{Episode Construction}
\dname{} contains 24 industrial scenes, of which 10 are designated as \emph{seen} scenes for training data collection and the remaining 14 are kept completely unseen during training to evaluate scene-level generalization.
For each seen scene, we sample 1200 training episodes and 300 test episodes.
For each unseen scene, we sample 300 test episodes only.
This results in 12000 training episodes, 3000 test episodes on seen scenes, and 4200 test episodes on unseen scenes.
We use the same sampling protocol for all four benchmark difficulty levels, ensuring that performance differences across acoustic conditions are not confounded by different start--goal distributions.
Across difficulty levels, the underlying scene split and episode configurations are kept consistent.
This design allows us to separately evaluate in-distribution navigation on seen environments and generalization to entirely unseen industrial layouts.
Additional details are provided in App.~\ref{app:benchmark}.

\section{Method}
\label{sec:method}
\mname{} is trained in three stages (Fig.~\ref{fig:overview}).
First, we augment the audio pathway with a lightweight spatial-audio adapter (SAA) to equip the model with directional acoustic perception.
Second, we perform multimodal supervised fine-tuning (SFT) on audio-depth navigation trajectories, enabling the model to jointly reason over global cues from spatial audio and local geometric information from depth.
Third, we extend the supervised policy into a backbone-shared actor--critic architecture, where the original action output serves as the actor and an additional critic head estimates the state value.
We optimize the policy with proximal policy optimization (PPO) using our proposed TARS.

\subsection{Spatial Audio Adaptation}
\label{subsec:adaptation}
The audio pathways of existing LMMs are typically trained for mono audio and not optimized to recover directional information from FOA signals.
This capability is particularly important in our setting, where the acoustic target provides the long-range cue for navigation.
We therefore augment the audio encoder with an SAA from SPUR~\cite{sakshi2025spur} and perform a spatial adaptation.  

The adapted audio representation is written as
\begin{equation}
    \mathbf{H}^{a}_t
    =
    E_{\mathrm{audio}}
    \left(
        \mathbf{X}^{a}_t;\phi
    \right),
    \label{eq:spatial_audio_encoder}
\end{equation}
where $E_{\mathrm{audio}}$ denotes the audio encoder augmented with SAA and $\phi$ denotes the SAA parameters.

To explicitly ground the learned audio representation in source direction, we attach a lightweight auxiliary regression head $g_{\mathrm{dir}}$ to the output of the audio encoder.
The encoder representation is first temporally aggregated and then mapped to an estimate of the relative source azimuth:
\begin{equation}
    \hat{\theta}_t
    =
    g_{\mathrm{dir}}
    \left(
        \operatorname{Pool}
        \left(
            \mathbf{H}^{a}_t
        \right)
    \right),
    \label{eq:direction_head}
\end{equation}
where $\hat{\theta}_t$ denotes the predicted agent-centric direction of arrival (DoA).
The adaptation objective is formulated as
\begin{equation}
    \mathcal{L}_{\mathrm{dir}}
    =
    \frac{1}{N}
    \sum_{t=1}^{N}
    \ell_{\mathrm{dir}}
    \left(
        \hat{\theta}_t,\theta_t
    \right),
    \label{eq:direction_loss}
\end{equation}
where $\theta_t$ is the derived ground-truth azimuth, and $\ell_{\mathrm{dir}}(\cdot)$ denotes the angular regression loss.

\subsection{Multimodal Supervised Fine-Tuning}
\label{subsec:sft}

We then use SFT to initialize the policy from expert navigation demonstrations. 
Each demonstration is converted into a sequence of state-action pairs
\(\{(o_t,a_t^\star)\}_{t=1}^{T}\), where \(o_t\) contains the agent's multimodal observation and \(a_t^\star\) is the expert action from an oracle trajectory planner.
Given the current multimodal observation, the model predicts
the next discrete action as a short text response.

Formally, let \(x_t\) denote the multimodal input sequence constructed from the
observation \(o_t\), and let \(y_t\) be the textual tokenization of the expert
action \(a_t^\star\). SFT minimizes the teacher-forced negative log-likelihood
of the expert action:
\[
\mathcal{L}_{\mathrm{sft}}
=
-\mathbb{E}_{(o_t,a_t^\star)\sim\mathcal{D}}
\left[
\sum_{k=1}^{|y_t|}
\log p_\theta(y_{t,k}\mid x_t, y_{t,<k})
\right].
\]
Only the action-answer tokens contribute to the loss.
The resulting SFT policy serves as the initialization for subsequent policy optimization.

\subsection{Reinforcement Learning with Transition-Aware Reward Shaping}
\label{subsec:rl}
After SFT, we further optimize the policy with PPO to improve closed-loop
behavior under states induced by the policy itself.
\paragraph{Backbone-Shared Actor--Critic}
To avoid duplicating the large multimodal backbone for PPO, we use a backbone-shared but gradient-decoupled actor--critic. Both branches reuse the final-token multimodal representation $\mathbf{h}_t$. The actor constructs a categorical policy over the valid navigation tokens. 
The critic predicts the state value from $\operatorname{stopgrad}(\mathbf{h}_t)$ observing the normalized remaining episode budget.
Full architectural and optimization details are provided in App.~\ref{app:actor-critic}.

\paragraph{Transition-Aware Reward Shaping}
Prior audio-guided navigation methods typically shape rewards using geodesic progress toward the goal~\cite{chen2020soundspaces,zeng2026semantic}.
While geodesic distance accounts for obstacle geometry, it depends only on the agent's position and is insensitive to its heading.
Consequently, in-place rotations produce zero geodesic progress, regardless of whether they align the agent with a feasible route or move it away from one.
This leads to ambiguous credit assignment for discrete navigation actions.

To address this limitation, we introduce TARS, which defines a pose-aware action-distance potential over a position--heading graph.
The graph captures spatial feasibility and orientation-dependent navigation costs under the agent's discrete motion primitives.
Let $N(s)$ denote the minimum number of executable actions required to reach the goal region from state $s$, accounting for both position and heading.
Unlike geodesic distance, $N(s)$ directly measures the remaining navigation effort in the policy's action space.

Building on potential-based reward shaping~\cite{ng1999policy}, we define
\begin{equation}
\Phi(s) = -N(s),
\end{equation}
and construct the transition reward as
\begin{equation}
r_{\mathrm{TARS}}(s_t,s_{t+1})
= \gamma_\Phi \Phi(s_{t+1})-\Phi(s_t),
\label{eq:tars}
\end{equation}
where $\gamma_\Phi=1$, such that the reward directly measures the reduction in remaining action count.
Unlike position-only geodesic progress, TARS distinguishes beneficial from counterproductive rotations, providing orientation-sensitive feedback even when the agent remains at the same physical location.
This aligns local reward feedback with long-horizon navigation progress in the agent's discrete action space.
The overall RL reward combines TARS with a per-step slack penalty $r_{\mathrm{slack}}$ and a terminal success reward:
\begin{equation}
r_t
=
r_{\mathrm{slack}}
+
r_{\mathrm{TARS}}(s_t,s_{t+1})
+
\mathbbm{1}_{\mathrm{succ}} r_{\mathrm{succ}},
\label{eq:rl_reward}
\end{equation}
where $r_{\mathrm{slack}}<0$ discourages unnecessarily long trajectories.
Upon success, we additionally reward path efficiency:
\begin{equation}
r_{\mathrm{succ}}
=
R_{\mathrm{succ}}\cdot \mathrm{SPL},
\qquad
\mathrm{SPL}
=
\frac{L^\star}{\max(L^\star,L)}.
\end{equation}
Here, $L^\star$ is the shortest-path length from the episode start to the goal, and $L$ is the traveled path length.
Under our strict stopping protocol, success requires the agent to issue \texttt{stop} within the target region:
\begin{equation}
\mathrm{succ}
=
\mathbbm{1}\{a_t=\texttt{stop}\}
\mathbbm{1}\{d_{\mathrm{euc}}(s_t,g)\le r_{\mathrm{goal}}\}.
\end{equation}

The action graph and privileged simulator state are used only for reward construction during training and are never exposed to the policy.
Graph construction and PPO implementation details are provided in App.~\ref{app:action_potential} and App.~\ref{app:ppo}, respectively.

\begin{table*}[t]
\centering
\caption{Navigation performance in seen scenes.}
\vspace{-.8em}
\label{tab:seen_navigation}
\resizebox{\textwidth}{!}{%
\begin{tabular}{l*{16}{c}}
\toprule
\toprule
\multirow{2}{*}{Method}
& \multicolumn{4}{c}{\codebf{Clean}}
& \multicolumn{4}{c}{\codebf{Prop}}
& \multicolumn{4}{c}{\codebf{DN}}
& \multicolumn{4}{c}{\codebf{LN}} \\
\cmidrule(lr){2-5}
\cmidrule(lr){6-9}
\cmidrule(lr){10-13}
\cmidrule(lr){14-17}
& SR$\uparrow$ & SPL$\uparrow$ & SNA$\uparrow$ & FDG$\downarrow$
& SR$\uparrow$ & SPL$\uparrow$ & SNA$\uparrow$ & FDG$\downarrow$
& SR$\uparrow$ & SPL$\uparrow$ & SNA$\uparrow$ & FDG$\downarrow$
& SR$\uparrow$ & SPL$\uparrow$ & SNA$\uparrow$ & FDG$\downarrow$ \\
\midrule

\multicolumn{17}{l}{\textbf{1. Oracle-assisted methods}} \\
\multicolumn{17}{l}{\quad {1.1 Zero-shot (strict stopping)}} \\

\rowcolor{uciblue!5}
Qwen 2.5-Omni~\cite{qwen2025qwen2.5}
& 19.00 & 13.60 & 7.70 & 15.19
& 16.00 & 13.00 & 6.70 & 14.96
& 22.00 & 17.29 & 9.35 & 14.11
& 19.37 & 13.91 & 8.10 & 15.64 \\

\rowcolor{uciblue!10}
Qwen 3-Omni~\cite{xu2025qwen3}
& 72.00 & 71.68 & 59.48 & 6.00
& 73.00 & 72.76 & 60.87 & 5.18
& 70.00 & 69.83 & 58.02 & 5.50
& 70.20 & 69.78 & 59.31 & 6.53 \\

\midrule
\multicolumn{17}{l}{\textbf{2. Non-oracle methods}} \\
\multicolumn{17}{l}{\quad {2.1 Zero-shot (relaxed stopping)}} \\

\rowcolor{uciblue!5}
Qwen 2.5-Omni~\cite{qwen2025qwen2.5}
& 1.67 & 1.67 & 1.56 & 24.66
& 2.67 & 2.36 & 1.74 & 24.39
& \underline{2.33} & 1.64 & 1.49 & \underline{24.29}
& 2.67 & 1.96 & \underline{1.85} & 24.35 \\

\rowcolor{uciblue!10}
Qwen 3-Omni~\cite{xu2025qwen3}
& 2.67 & 2.34 & 1.80 & 26.42
& 3.67 & 2.76 & 2.30 & 25.29
& \underline{2.33} & \underline{1.88} & \underline{1.67} & 26.44
& 2.67 & 2.13 & 1.59 & 25.71 \\

\addlinespace[2pt]
\multicolumn{17}{l}{\quad {2.2 Training-based (strict stopping)}} \\

\rowcolor{uciblue!5}
MAGNet~\cite{zeng2026semantic}
& \underline{92.23} & \underline{84.26} & \underline{68.48} & \underline{1.64}
& \underline{50.53} & \underline{38.90} & 16.43 & \underline{7.11}
& 0.00 & 0.00 & 0.00 & 16.73
& 1.73 & 1.33 & 1.76 & \underline{14.61} \\

\rowcolor{uciblue!10}
SAVi~\cite{chen2021semantic}
& 76.77 & 70.45 & 53.07 & 3.18
& 23.50 & 22.42 & 14.58 & 12.07
& 0.00 & 0.00 & 0.00 & 18.24
& 0.07 & 0.06 & 0.04 & 25.13 \\

\rowcolor{uciblue!5}
SoundSpaces~\cite{chen2020soundspaces}
& 74.47 & 69.57 & 53.50 & 4.72
& 40.27 & 30.68 & \underline{18.64} & 8.40
& 0.10 & 0.09 & 0.09 & 15.5
& \underline{2.80} & \underline{2.30} & 1.76 & \underline{14.61} \\

\rowcolor{uciblue!20}
\mname
& \textcolor{blue}{\textbf{97.70}} & \textcolor{blue}{\textbf{93.48}} & \textcolor{blue}{\textbf{72.74}} & \textcolor{blue}{\textbf{1.08}}
& \textcolor{blue}{\textbf{91.33}} & \textcolor{blue}{\textbf{82.14}} & \textcolor{blue}{\textbf{62.80}} & \textcolor{blue}{\textbf{2.01}}
& \textcolor{blue}{\textbf{69.70}} & \textcolor{blue}{\textbf{58.50}} & \textcolor{blue}{\textbf{40.00}} & \textcolor{blue}{\textbf{7.11}}
& \textcolor{blue}{\textbf{67.40}} & \textcolor{blue}{\textbf{56.51}} & \textcolor{blue}{\textbf{38.22}} & \textcolor{blue}{\textbf{7.54}} \\

\bottomrule
\bottomrule
\end{tabular}%
}
\begin{minipage}{\textwidth}
\footnotesize
\textit{Note:}
Oracle methods receive ground-truth DoA and source distance
(see Prompt~\ref{prompt:oracle}), whereas non-oracle methods
receive no ground-truth navigation information and use only
their permitted observations and action histories
(see Prompt~\ref{prompt:no-oracle}).
Non-oracle zero-shot methods use relaxed stopping, where
\texttt{stop} is disabled and success is registered upon
entering the target radius $\delta$.
All other methods follow the strict stopping criterion.
\textcolor{blue}{\textbf{Blue bold}} and \underline{underlined}
values indicate the best and second-best results among
non-oracle methods, respectively. Ties are marked equally.
\end{minipage}
\vspace{-1em}
\end{table*}

\begin{table*}[h]
\centering
\caption{Navigation performance in unseen scenes.}
\vspace{-.8em}
\label{tab:unseen_navigation}
\resizebox{\textwidth}{!}{%
\begin{tabular}{l*{16}{c}}
\toprule
\toprule
\multirow{2}{*}{Method}
& \multicolumn{4}{c}{\codebf{Clean}}
& \multicolumn{4}{c}{\codebf{Prop}}
& \multicolumn{4}{c}{\codebf{DN}}
& \multicolumn{4}{c}{\codebf{LN}} \\
\cmidrule(lr){2-5}
\cmidrule(lr){6-9}
\cmidrule(lr){10-13}
\cmidrule(lr){14-17}
& SR$\uparrow$ & SPL$\uparrow$ & SNA$\uparrow$ & FDG$\downarrow$
& SR$\uparrow$ & SPL$\uparrow$ & SNA$\uparrow$ & FDG$\downarrow$
& SR$\uparrow$ & SPL$\uparrow$ & SNA$\uparrow$ & FDG$\downarrow$
& SR$\uparrow$ & SPL$\uparrow$ & SNA$\uparrow$ & FDG$\downarrow$ \\
\midrule

\multicolumn{17}{l}{\textbf{1. Oracle-assisted methods (zero-shot)}} \\
\multicolumn{17}{l}{\quad {1.1 Zero-shot (strict stopping)}} \\

\rowcolor{uciblue!5}
Qwen 2.5-Omni~\cite{qwen2025qwen2.5}
& 14.29 & 11.40 & 6.12 & 16.63
& 12.71 & 10.06 & 5.55 & 16.79
& 15.10 & 12.20 & 6.41 & 16.25
& 17.10 & 14.34 & 7.88 & 16.30 \\

\rowcolor{uciblue!10}
Qwen 3-Omni~\cite{xu2025qwen3}
& 62.14 & 61.79 & 52.12 & 6.36
& 62.38 & 61.93 & 52.87 & 6.40
& 63.33 & 62.92 & 53.27 & 6.16
& 62.62 & 62.24 & 52.70 & 6.31 \\

\midrule
\multicolumn{17}{l}{\textbf{2. Non-oracle methods}} \\
\multicolumn{17}{l}{\quad {2.1 Zero-shot (relaxed stopping)}} \\

\rowcolor{uciblue!5}
Qwen 2.5-Omni~\cite{qwen2025qwen2.5}
& 2.38 & 2.34 & 2.12 & 23.60
& 2.14 & 2.11 & 1.96 & 23.33
& 2.62 & 2.59 & 2.37 & 23.44
& 2.86 & 2.73 & 2.46 & 23.74 \\

\rowcolor{uciblue!10}
Qwen 3-Omni~\cite{xu2025qwen3}
& 4.29 & 3.70 & 2.77 & 24.99
& 2.86 & 2.62 & 1.97 & 25.00
& \underline{4.29} & \underline{3.72} & \underline{2.59} & 24.69
& \underline{5.00} & \underline{4.03} & \underline{2.78} & 24.72 \\

\addlinespace[2pt]
\multicolumn{17}{l}{\quad {2.2 Training-based (strict stopping)}} \\

\rowcolor{uciblue!5}
MAGNet~\cite{zeng2026semantic}
& \underline{85.60} & \underline{79.20} & \underline{64.22} & \underline{2.73}
& \underline{40.71} & \underline{30.30} & 12.66 & \underline{9.24}
& 0.00 & 0.00 & 0.00 & 17.59
& 1.71 & 1.23 & 1.01 & 16.77 \\

\rowcolor{uciblue!10}
SAVi~\cite{chen2021semantic}
& 72.31 & 67.10 & 51.36 & 4.03
& 17.40 & 16.30 & 10.01 & 13.00
& 0.00 & 0.00 & 0.00 & 18.17
& 0.14 & 0.13 & 0.11 & 22.41 \\

\rowcolor{uciblue!5}
SoundSpaces~\cite{chen2020soundspaces}
& 71.00 & 67.20 & 51.90 & 5.09
& 34.40 & 27.20 & \underline{16.20} & 10.10
& 0.29 & 0.18 & 0.14 & \underline{15.92}
& 2.62 & 2.11 & 1.63 & \underline{14.93} \\

\rowcolor{uciblue!20}
\mname
& \textcolor{blue}{\textbf{90.60}} & \textcolor{blue}{\textbf{86.00}} & \textcolor{blue}{\textbf{66.07}} & \textcolor{blue}{\textbf{2.27}}
& \textcolor{blue}{\textbf{81.60}} & \textcolor{blue}{\textbf{72.80}} & \textcolor{blue}{\textbf{54.86}} & \textcolor{blue}{\textbf{4.63}}
& \textcolor{blue}{\textbf{62.00}} & \textcolor{blue}{\textbf{51.70}} & \textcolor{blue}{\textbf{33.87}} & \textcolor{blue}{\textbf{9.60}}
& \textcolor{blue}{\textbf{60.00}} & \textcolor{blue}{\textbf{49.00}} & \textcolor{blue}{\textbf{32.05}} & \textcolor{blue}{\textbf{8.96}} \\

\bottomrule
\bottomrule
\end{tabular}%
}
\begin{minipage}{\textwidth}
\footnotesize
\textit{Note:}
We use the same prompt and stopping protocols as in
Tab.~\ref{tab:seen_navigation}.
\textcolor{blue}{\textbf{Blue bold}} and \underline{underlined}
values indicate the best and second-best results among
non-oracle methods only, respectively.
\end{minipage}
\vspace{-2.2em}
\end{table*}

\subsection{Bounded Action History}
In addition to the current audio-depth observation, \mname{} optionally conditions each decision on a short history of the eight most recent
navigation actions. The history is serialized directly as part of the
textual prompt and therefore introduces no additional trainable parameters. At timestep $t$, we denote this context as
\[
\mathcal{M}_t =
(a_{\max(0,t-8)},\ldots,a_{t-1}).
\]
During closed-loop interaction, $\mathcal{M}_t$ is updated after each
executed action. The same interface is used during training and inference.

\section{Experimental Results}
\label{sec:results}

\paragraph{Evaluation Metrics}
Following prior studies~\cite{chen2021semantic}, we evaluate navigation performance using four metrics: success rate (SR), success weighted by path length (SPL), success weighted by number of actions (SNA), and final distance to goal (FDG).
Lower values indicate better performance for FDG, and higher values indicate better performance for the other metrics.
Detailed definitions are provided in App.~\ref{app:metrics}.
\vspace{-.5em}
\paragraph{Implementation Details}
All models are built on Qwen2.5-Omni-7B~\cite{qwen2025qwen2.5} and trained on NVIDIA H200 GPUs.
Detailed optimization settings, rollout configuration, critic warm-up, and SFT anchoring are provided in App.~\ref{app:impl_details}.

\paragraph{Baselines}
We compare \mname{} against multiple baselines across two paradigms.
For \textbf{trained navigation methods}, we consider
MAGNet~\cite{zeng2026semantic},
SAVi~\cite{chen2021semantic}, and
SoundSpaces~\cite{chen2020soundspaces}.
For \textbf{zero-shot LMM agents}, we evaluate native multimodal models,
including Qwen 2.5-Omni~\cite{qwen2025qwen2.5},
Qwen 3-Omni~\cite{xu2025qwen3}.
For zero-shot LMMs without oracle information, we adopt a relaxed stopping protocol
that automatically registers success upon entering the goal radius, since
these models are not provided with a reliable metric source-distance cue.
This relaxation favors the zero-shot baselines and prevents their performance from being dominated by an ill-posed stopping decision based on acoustic amplitude.
Additional details are provided in App.~\ref{app:zero_shot_baselines}.

\subsection{Main Results}
\begin{figure*}[tb]
\centering
\resizebox{\textwidth}{!}{
\includegraphics{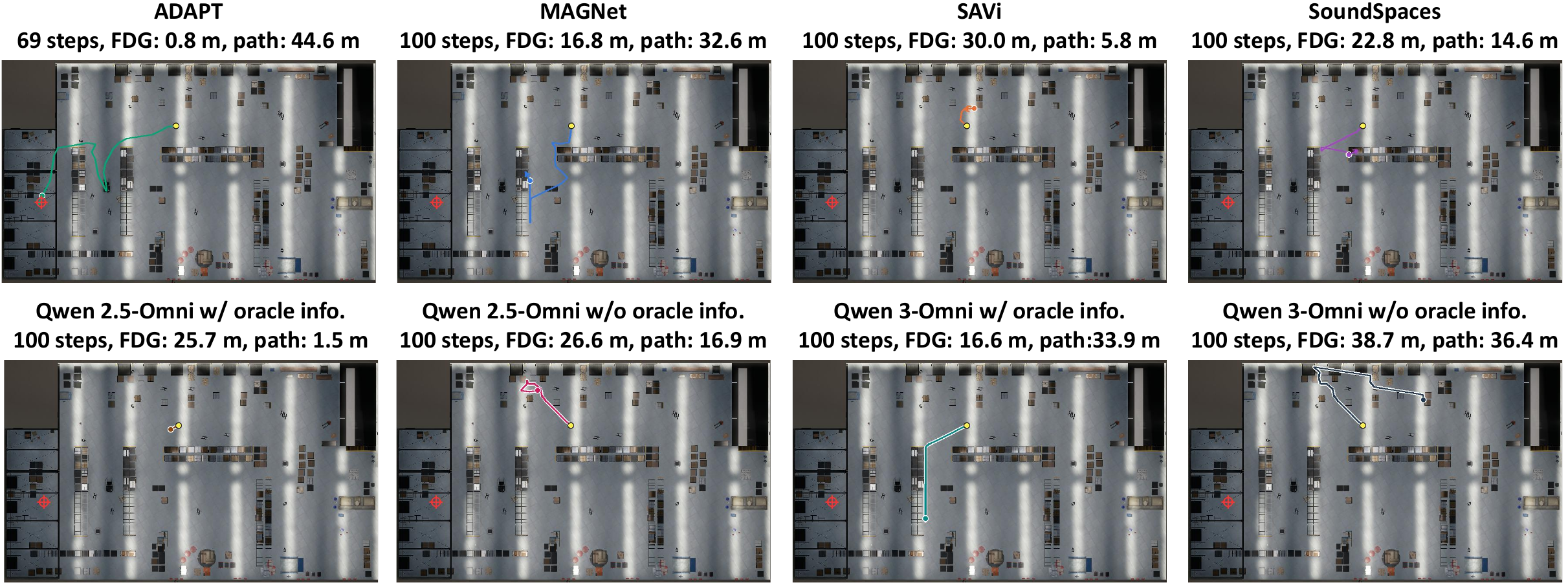}
}
\vspace{-2em}
\caption{\textbf{Qualitative comparison of navigation trajectories.}
Each panel reports the number of executed actions, FDG, and traveled path length. More examples are included in the appendix (see Fig.~\ref{fig:app_qualitative}).
}
\vspace{-1.5em}
\label{fig:qualitative}
\end{figure*}

Tabs.~\ref{tab:seen_navigation} and~\ref{tab:unseen_navigation} show that \mname{} consistently outperforms training-based baselines and non-oracle zero-shot LMMs across both seen and unseen scenes. 
The advantage becomes more obvious under increasingly challenging acoustic conditions, where the baselines degrade substantially while \mname{} remains effective. 
Performance also transfers well to unseen scenes, indicating that the learned policy generalizes beyond scene-specific geometry and acoustic patterns.
Meanwhile, the large gap between non-oracle and oracle-assisted zero-shot LMMs suggests that reliable spatial-audio perception remains a major bottleneck for general-purpose models, while successful navigation further requires effective integration of acoustic guidance with local geometric information.

As shown in Fig.~\ref{fig:qualitative}, \mname{} maintains consistent goal-directed progress and reaches the target, while competing methods either stagnate or deviate from an effective route.
The short trajectories of some failed baselines reflect limited progress rather than navigation efficiency.
Moreover, even oracle-assisted zero-shot LMMs may fail to translate accurate target information into effective obstacle-aware closed-loop behavior, further motivating task-specific audio-depth adaptation and policy optimization.

\subsection{Discussion}
Tab.~\ref{tab:ablation} demonstrates the complementary roles of SAA and TARS in our framework. SAA consistently improves the policy during SFT, suggesting that it enables more effective adaptation to audio-depth observations. 
The RL ablation compares TARS against geodesic-distance-based reward shaping under matched SAA and action-history configurations. TARS improves navigation performance across the evaluated configurations, with the largest gains observed when action history is absent. This suggests that pose-aware action-distance feedback provides useful supervision beyond position-based geodesic progress, particularly for learning effective rotational decisions.
Combining SAA and TARS yields clear gains over either component alone, showing that representation adaptation and task-aware reward optimization address different aspects of the problem. 
Conditioning on a short action history further improves robustness by
providing lightweight temporal context without introducing an additional
memory module.

\begin{table}[t]
\centering
\begin{threeparttable}

\caption{Ablation study under the \codebf{LN} setting.
(a) SFT-only ablation. (b) RL ablation.}
\vspace{-.1em}
\label{tab:ablation}

\footnotesize
\setlength{\tabcolsep}{4pt}

\begin{tabular}{@{}c@{}}
\textbf{(a) SFT-only ablation} \\[2pt]

\resizebox{0.8\columnwidth}{!}{%
\begin{tabular}{cccccc}
\toprule
\toprule
\multicolumn{2}{c}{Component}
& \multicolumn{4}{c}{\codebf{LN}} \\
\cmidrule(lr){1-2}
\cmidrule(lr){3-6}
SAA & Action Hist.\tnote{*}
& SR$\uparrow$ & SPL$\uparrow$ & SNA$\uparrow$ & FDG$\downarrow$ \\
\midrule

\rowcolor{uciblue!5}
$\times$ & $\times$
& 52.07 & 42.86 & 30.84 & 12.59 \\

\rowcolor{uciblue!10}
$\times$ & $\checkmark$
& 58.73 & 47.74 & 33.13 & 10.00 \\

\rowcolor{uciblue!5}
$\checkmark$ & $\times$
& 53.27 & 46.29 & 31.89 & 12.15 \\

\rowcolor{uciblue!10}
$\checkmark$ & $\checkmark$
& 62.93 & 52.58 & 35.42 & 9.00 \\

\bottomrule
\bottomrule
\end{tabular}%
}
\\[2em]

\textbf{(b) RL ablation} \\[2pt]

\resizebox{0.8\columnwidth}{!}{%
\begin{tabular}{ccccccc}
\toprule
\toprule
\multicolumn{3}{c}{Component}
& \multicolumn{4}{c}{\codebf{LN}} \\
\cmidrule(lr){1-3}
\cmidrule(lr){4-7}
SAA & TARS\tnote{**} & Action Hist.
& SR$\uparrow$ & SPL$\uparrow$ & SNA$\uparrow$ & FDG$\downarrow$ \\
\midrule

\rowcolor{uciblue!5}
$\times$ & $\times$ & $\times$
& 53.93 & 46.28 & 33.26 & 12.14 \\

\rowcolor{uciblue!10}
$\times$ & $\checkmark$ & $\times$
& 57.67 & 48.43 & 35.13 & 11.66 \\

\rowcolor{uciblue!5}
$\times$ & $\times$ & $\checkmark$
& 65.93 & 53.77 & 37.85 & 8.59 \\

\rowcolor{uciblue!10}
$\times$ & $\checkmark$ & $\checkmark$
& 66.87 & 55.50 & 39.58 & 7.65 \\

\rowcolor{uciblue!5}
$\checkmark$ & $\times$ & $\times$
& 56.60 & 50.05 & 35.04 & 12.00 \\

\rowcolor{uciblue!10}
$\checkmark$ & $\checkmark$ & $\times$
& 62.87 & 52.75 & 38.98 & 10.52 \\

\rowcolor{uciblue!5}
$\checkmark$ & $\times$ & $\checkmark$
& 66.43 & 56.14 & 38.38 & 7.67 \\

\rowcolor{uciblue!10}
$\checkmark$ & $\checkmark$ & $\checkmark$
& 67.40 & 56.51 & 38.22 & 7.54 \\

\bottomrule
\bottomrule
\end{tabular}%
}

\end{tabular}

\begin{tablenotes}[flushleft]
\footnotesize
\item[*] Action Hist. denotes conditioning the policy on the eight most recent navigation actions.
\item[**] $\times$ denotes geodesic-distance reward shaping.
\end{tablenotes}

\end{threeparttable}
\vspace{-2.5em}
\end{table}

Acoustic difficulty induces clear yet non-uniform effects on DoA estimation, as presented in Tab.~\ref{tab:acoustic_conditions} and Fig.~\ref{fig:doa_acc} (in appendix). Under \codebf{DN}, performance degrades monotonically as SNR decreases from $20\text{ dB}$ to $-10\text{ dB}$, showing that stronger interference primarily impairs fine-grained DoA estimation while coarse-grained alignment remains relatively resilient.
Counterintuitively, \mname{} under \codebf{Prop} yields the best overall estimation, even outperforming the results under \codebf{Clean}.
This inversion highlights an acoustic-domain bias: the noise-trained SAA is better calibrated to perturbed or propagated sound fields than to idealized direct-path signals. 
Furthermore, across all difficulties, the substantial gap between median error ($\sim10^\circ$) and MAE ($>25^\circ$) uncovers a heavy-tailed failure mode. While most predictions are highly accurate, a subset of catastrophic directional errors disproportionately inflates the mean error. Overall, acoustic difficulty impacts not only average precision, but also domain calibration and failure dynamics.

Tab.~\ref{tab:dn_snr} shows a clear degradation in navigation performance as the diffuse-noise level increases. At relatively high SNR, the agent maintains strong success and path efficiency, indicating that the audio cue remains sufficiently informative for reliable navigation. As SNR decreases, SR, SPL, and SNA all drop consistently, while FDG increases. 
The performance decline becomes particularly obvious at the lowest SNR, suggesting that severe diffuse noise substantially corrupts the directional information available from the FOA signal. Overall, the results demonstrate reasonable robustness under moderate noise, while also revealing the expected limitation under extremely low-SNR conditions.

\begin{table}[t]
\centering
\caption{DoA estimation under different difficulties.}
\vspace{-.7em}
\label{tab:acoustic_conditions}
\resizebox{\columnwidth}{!}{%
\begin{tabular}{lcccccc}
\toprule
\toprule
Metric & \codebf{Clean} & \codebf{Prop}
& \codebf{DN}@$20$dB & \codebf{DN}@$10$dB
& \codebf{DN}@$0$dB & \codebf{DN}@$-10$dB \\
\midrule
\rowcolor{uciblue!5}
MAE $\downarrow$
& $35.45^\circ$
& \textcolor{blue}{$\mathbf{25.03^\circ}$}
& $\underline{28.45^\circ}$
& $32.02^\circ$
& $34.06^\circ$
& $34.17^\circ$ \\
\rowcolor{uciblue!10}
Median error $\downarrow$
& $13.23^\circ$
& $\underline{10.93^\circ}$
& \textcolor{blue}{$\mathbf{10.74^\circ}$}
& $11.47^\circ$
& $14.93^\circ$
& $17.89^\circ$ \\
\bottomrule
\bottomrule
\end{tabular}%
}
\vspace{-1em}
\end{table}

\begin{table}[t]
\centering
\caption{Navigation performance under different SNR under the \codebf{DN} setting.}
\vspace{-.8em}
\label{tab:dn_snr}
\resizebox{.8\columnwidth}{!}{%
\begin{tabular}{ccccc}
\toprule
\toprule
SNR (dB) & SR$\uparrow$ & SPL$\uparrow$ & SNA$\uparrow$ & FDG$\downarrow$ \\
\midrule
\rowcolor{uciblue!5}
10  & 67.10 & 59.30 & 39.00 & 7.89  \\
\rowcolor{uciblue!10}
0   & 61.90 & 51.00 & 32.60 & 9.18  \\
\rowcolor{uciblue!5}
-5  & 52.60 & 41.50 & 27.40 & 12.42 \\
\rowcolor{uciblue!10}
-10 & 31.20 & 23.70 & 14.40 & 18.39 \\
\bottomrule
\bottomrule
\end{tabular}%
}
\begin{minipage}{\columnwidth}
\footnotesize
\textit{Note:} 
Unlike the main \codebf{DN} evaluation, where the nominal SNR is randomized
episode-wise, this analysis fixes the SNR to specific values in order
to isolate the effect of noise strength.
\end{minipage}
\vspace{-1.8em}
\end{table}

\section{Related Work}
\paragraph{Vision-Language-Action Models}
Recent VLA models bridge pretrained vision-language representations with embodied control by directly predicting robot actions from multimodal observations and language instructions.
Representative models such as RT-2~\cite{brohan2023rt} formulate actions as tokens within a vision-language model, while OpenVLA~\cite{kim2024openvla} further demonstrates scalable open-source VLA training across diverse robot demonstrations.
However, existing VLA models predominantly rely on RGB observations and are largely developed for visually grounded robotic manipulation.
Our work extends this action-generation paradigm to audio-depth perception, enabling embodied action planning when reliable visual observations are unavailable.

\paragraph{Audio-Visual Navigation}
Audio-visual navigation extends embodied navigation by incorporating spatial audio as an additional cue for goal localization.
SoundSpaces~\cite{chen2020soundspaces} established a widely used simulation framework in which agents jointly exploit visual observations and binaural audio to navigate toward sound-emitting targets.
Subsequent works improve audio-visual reasoning through structured acoustic memory and learned waypoints~\cite{chen2020learning}, or consider semantic and temporally sparse acoustic events~\cite{chen2021semantic}.
These approaches demonstrate the value of spatial audio for navigation, but largely retain vision as a primary perceptual modality.
In contrast, we study navigation under \emph{lights-out} conditions, where RGB perception is unavailable, and combine spatial audio with depth sensing for goal-directed navigation and collision avoidance.

\paragraph{Acoustic-Guided Navigation and Active Acoustic Sensing}
Acoustic cues have also been used directly for robotic localization and navigation.
Early sound-source-seeking systems equipped mobile robots with microphone arrays to estimate the direction of an acoustic target and move toward it, while acoustic-field navigation methods construct designed sound fields or acoustic trails that guide robots toward specified destinations~\cite{tsunoda2019experimental}.
Modern audio-goal methods similarly treat a continuously emitting sound source as the navigation target~\cite{chen2020soundspaces,yu2022sound,kondoh2023multi}, but typically combine this acoustic guidance with RGB-based perception.
A separate line of work investigates active acoustic sensing.
For example, VisualEchoes~\cite{gao2020visualechoes} actively emits acoustic chirps and analyzes their echoes to infer spatial properties of the environment.
In contrast to both RGB-dependent audio-goal navigation and echo-based scene sensing, \mname{} uses spatial acoustic guidance to indicate the navigation objective while depth observations provide obstacle geometry, enabling direct action planning in lights-out industrial environments.

\vspace{-.5em}
\section{Conclusion}
\vspace{-.5em}
\label{sec:conclusion}
We introduced \dname{} and \mname{} for audio-depth navigation in lights-out industrial environments.
\mname{} combines spatial-audio adaptation with TARS, which measures navigation progress over a
position--heading action graph.
Experiments show that \mname{} consistently outperforms the baselines across seen and unseen scenes.
Ablations further confirm the benefits of SAA and TARS.
Overall, our results demonstrate the potential of spatial audio and depth for reliable RGB-free navigation in lights-out industrial environments.

\clearpage
\section*{Limitations}
Our study focuses on controlled, high-fidelity simulation rather than
physical robot deployment. \dname{} incorporates cluttered industrial
layouts, dynamic agents, geometry-dependent acoustic propagation, diffuse
background noise, and localized machinery sounds, which enables systematic
evaluation under acoustic conditions that are difficult to reproduce
consistently in the real world. Nevertheless, simulation cannot capture all
effects of physical microphones and depth sensors, such as channel mismatch,
calibration drift, sensor noise, or unmodeled acoustic dynamics. 
Evaluating \mname{} on a physical audio-depth robot is therefore an
important direction for future work.

\dname{} also uses a controlled acoustic beacon: the target
emits the same periodic chirp across scenes and episodes, while localized
equipment categories use representative sound loops. This design removes
semantic sound recognition as a confounding factor and allows the benchmark
to isolate spatial perception and closed-loop navigation. However, real
industrial environments may contain intermittent targets, previously unseen
signals, and distractors with stronger spectral or temporal similarity to
the navigation beacon. Future extensions can diversify target waveforms,
number of targets,
source levels, temporal patterns, and distractor characteristics, and expand
evaluation beyond the current environments.

Finally, expert demonstrations and TARS make use of privileged simulator
information during training to construct expert actions and action-space
progress signals. These variables are never provided to the deployed policy. The current formulation
therefore demonstrates that privileged training signals can be distilled
into an observation-driven navigation policy, but assumes access to such
signals during training. Extending the framework to map-free reward
estimation, continuous control, and online adaptation would broaden its
applicability. In addition, current general-purpose LMMs lack a native FOA
interface, requiring our zero-shot baselines to receive synchronized FOA
channels through separate mono-audio inputs. Their results should therefore
be interpreted as a diagnostic of current spatial-audio capability rather
than an upper bound on future spatially native LMMs.

\section*{Ethical Considerations}
This work investigates audio-depth navigation in lights-out industrial environments using simulation-based benchmarks. 
All navigation experiments are conducted in virtual environments without human participants or the intentional collection of personally identifiable information. Third-party audio recordings used for acoustic simulation are subject to their respective licensing and attribution requirements.
However, navigation performance in simulation does not guarantee safe operation in real-world factories, where complex acoustic conditions, dynamic obstacles, and unpredictable human activities may introduce additional risks. 
Deployment in physical environments would require extensive real-world validation. 
Acoustic sensing may also inadvertently capture sensitive operational information, necessitating appropriate privacy protections. 
We therefore present \mname{} as a research framework rather than a deployment-ready or safety-certified navigation system.


\bibliography{custom}


\clearpage
\appendix
\startcontents[appendices]
\section*{Contents of Appendix}
\printcontents[appendices]{}{1}{\normalsize}

\section{Task Formulation Details}
\label{app:task_formulation}
\paragraph{Environment State}
At timestep $t$, the underlying simulator state is
\begin{equation}
    S_t =
    \left(
        \mathbf{p}_t,\,
        \psi_t,\,
        \mathbf{p}^{s},\,
        \mathcal{G}_t
    \right),
\end{equation}
where $\mathbf{p}_t=(x_t,y_t)$ is the planar position of the robot, $\psi_t$ is its heading, $\mathbf{p}^{s}=(x^{s},y^{s})$ is the target sound-source position, and $\mathcal{G}_t$ represents the surrounding geometric configuration, including static or dynamic obstacles.
The simulator has access to $S_t$ for environment transitions and supervision generation, but the policy does not directly observe these privileged variables.

\paragraph{Direction of Arrival Calculation}
The global direction of the sound source is computed as
\begin{equation}
    \theta^{\mathrm{world}}_t =
    \operatorname{atan2}
    \left(
        y^{s}-y_t,\,
        x^{s}-x_t
    \right).
\end{equation}
We then transform it into the agent-centric reference frame:
\begin{equation}
    \theta_t =
    \operatorname{wrap}
    \left(
        \theta^{\mathrm{world}}_t-\psi_t
    \right),
    \qquad
    \theta_t \in [-\pi,\pi),
    \label{eq:relative_azimuth}
\end{equation}
where $\operatorname{wrap}(\cdot)$ maps the angular difference to the principal interval.
The resulting $\theta_t$ represents the relative direction of arrival (DoA) of the target sound with respect to the robot orientation.

\paragraph{Spatial-Audio and Depth Observations}
The spatial-audio observation is represented using four-channel first-order ambisonics (FOA):
\begin{equation}
    \mathbf{X}^{a}_t
    =
    \left[
        \mathbf{x}^{W}_t,\,
        \mathbf{x}^{Y}_t,\,
        \mathbf{x}^{Z}_t,\,
        \mathbf{x}^{X}_t
    \right]
    \in \mathbb{R}^{4 \times T_a},
\end{equation}
where $T_a$ denotes the number of audio samples in each observation window.
FOA signals are stored using the ACN/N3D convention with channel order $W$, $Y$ (left), $Z$ (up), and $X$ (front).
The FOA channels jointly encode both acoustic content and the directional structure of the surrounding sound field.
In our setting, the active target provides the primary acoustic cue for goal localization.

The egocentric depth observation is
\begin{equation}
    \mathbf{X}^{d}_t
    \in
    \mathbb{R}^{H \times W},
\end{equation}
where each pixel records the distance from the agent's depth sensor to the first visible surface along the corresponding viewing ray.
The two modalities play complementary roles: spatial audio indicates \emph{\textbf{where the target is}}, while depth indicates \emph{\textbf{where the agent can safely move}}.

\paragraph{Policy and Environment Transition}
At each timestep, the policy defines a categorical distribution over the discrete action space:
\begin{equation}
    \pi_{\theta}
    \left(
        a_t
        \mid
        O_{t}, a_{\leq t-1}
    \right)
    =
    \pi_{\theta}
    \left(
        a_t
        \mid
        \mathbf{X}^{a}_{t},
        \mathbf{X}^{d}_{t}, a_{\leq t - 1}
    \right),
\end{equation}
where $\theta$ denotes the policy parameters.
An action is sampled or selected according to
\begin{equation}
    a_t
    \sim
    \pi_{\theta}
    \left(
        \cdot \mid O_{t}, a_{\leq t-1}
    \right).
\end{equation}
After executing $a_t$, the simulator transitions according to
\begin{equation}
    S_{t+1}
    \sim
    \mathcal{T}
    \left(
        S_{t+1}
        \mid
        S_t,a_t
    \right),
    \label{eq:transition}
\end{equation}
and generates the next observation $O_{t+1}$.

\paragraph{Trajectory and Episode Termination}
A navigation trajectory of length $T$ is written as
\begin{equation}
    \tau =
    \left(
        O_0,a_0,
        O_1,a_1,
        \ldots,
        O_T
    \right).
\end{equation}
The agent is required to approach the acoustic target while avoiding collisions and unnecessary actions.
The episode terminates either when the robot reaches the target neighborhood and issues the stop action, or when the maximum number of navigation steps $T_{\max}$ is reached.

Although robot pose, target location, and environment geometry are available during simulation and may be used to construct training signals, \textbf{none of these privileged variables are provided to the deployed policy}.
Consequently, successful navigation must emerge from reasoning over spatial audio and depth observations alone.

\section{Benchmark Construction Details}
\label{app:benchmark}
\paragraph{IndustryNav Extension}
\dname{} is built upon IndustryNav~\cite{li2025industrynav}, which provides high-fidelity Unity warehouse environments containing diverse industrial assets, cluttered layouts, moving workers and equipment, collision-aware navigation, and synchronized egocentric sensing.
We retain the original navigation action space, motion configuration, scene geometry, dynamics, physics, and depth sensing,   while extending the environment
with spatial audio observations and acoustic scene modeling.
RGB frames may still be rendered for visualization and debugging, but they are never exposed to the agent.

\paragraph{Navigation Dynamics and Action Geometry}
We use a discrete navigation action space consisting of
\texttt{forward}, \texttt{left}, \texttt{right}, and \texttt{stop}.
A \texttt{forward} action moves the agent nominally $1.5\,\mathrm{m}$
along its current heading, while \texttt{left} and \texttt{right} rotate
the agent by $18^\circ$ in place in the corresponding direction.
The \texttt{stop} action terminates the episode.
The navigation target is considered reached when the agent is within a
radius of $\delta=1.5\,\mathrm{m}$ from the acoustic source.
Under the standard evaluation protocol, an episode is counted as successful
only when the agent issues \texttt{stop} within this target region;
otherwise, it terminates upon \texttt{stop} or after the maximum horizon
of $T_{\max}=100$ actions.
The same action geometry is used for demonstration collection, policy
training, and evaluation.

\paragraph{Acoustic Target Generation}
Each navigation target is associated with a stationary mono acoustic emitter located at the goal position.
The emitter continuously plays a fixed linear chirp sweeping from 500\,Hz to 900\,Hz over 50\,ms at constant amplitude.
The chirp is looped without gaps, corresponding to a repetition rate of 20\,Hz, and the same signal is used across all goals, scenes, and difficulty settings.
This controlled acoustic carrier encourages the agent to rely on spatial acoustic cues rather than semantic audio recognition.

We integrate a modified Steam Audio pipeline into the environment.
The target emitter is connected to the spatial renderer, while an FOA receiver is attached to the embodied agent.
As the agent moves and rotates, the renderer continuously updates the received sound field according to the relative source--listener configuration and geometry-dependent acoustic propagation.
Consequently, the acoustic observation varies jointly with the agent pose and the surrounding scene, forming a closed-loop sensory signal that evolves along the navigation trajectory.

\paragraph{Audio Synchronization and Observation Recording}
Audio is sampled at 48\,kHz using frames of 1024 samples, corresponding to 21.33\,ms per frame (46.875\,Hz).
The simulation timestep is synchronized with the audio frame duration such that each physics step corresponds to one audio frame, which makes the generated audio deterministic across execution modes.
At each navigation decision step, the agent remains stationary while a fixed temporal window of consecutive audio frames is recorded.
We use 24 frames per observation, resulting in
\[
T_a = 24 \times 1024 = 24576
\]
samples, corresponding to a 0.512\,s observation window.
For each decision step, we store the four-channel FOA waveform together with frame-level diagnostics, including per-channel RMS energy and the ground-truth DoA.

\paragraph{Acoustic Scene Modeling}
To model geometry-dependent propagation, all active static scene meshes are associated with acoustic material properties and merged into a scene-level static acoustic mesh.
Each acoustic material is characterized by frequency-dependent absorption and transmission coefficients, together with a scattering coefficient.
Specifically, \emph{absorption} measures the fraction of incident acoustic energy absorbed by a surface, while \emph{transmission} controls the fraction of sound energy propagating through the surface.
Both are specified separately for low-, mid-, and high-frequency bands.
The \emph{scattering} coefficient controls how strongly reflected sound is redistributed diffusely rather than following purely specular reflection.
Higher absorption therefore produces weaker reflections, higher transmission allows more sound to propagate through an object, and higher scattering yields more diffuse reflected sound.

Distance attenuation and air absorption are enabled in all benchmark settings, while geometry-dependent propagation consists of occlusion, frequency-dependent transmission, and multi-bounce reflections.
We use volumetric occlusion with a 0.5 m source radius and 8 samples, and simulate transmission using frequency-dependent material coefficients.
Reflections are computed with Steam Audio's CPU ray tracer using 512 rays, 8 bounces, 1024 diffuse samples, and a 1 s impulse response.
These propagation effects define the acoustic difficulty hierarchy used in our benchmark.
The \codebf{Clean} setting disables occlusion, transmission, and reflections, retaining only the direct-path signal with distance and air attenuation.
The \codebf{Prop} setting enables all geometry-dependent propagation effects.
\codebf{DN} builds on \codebf{Prop} by additionally introducing the diffuse background-noise field, while \codebf{LN} further incorporates sounds emitted by selected local scene objects.

We define five acoustic presets: \codebf{Concrete}, \codebf{Metal}, \codebf{Wood}, \codebf{Plastic}, and \codebf{Rubber}. The \codebf{Concrete}, \codebf{Metal}, and \codebf{Wood} parameters are taken directly from the material library distributed with Steam Audio 4.8.1, whereas \codebf{Plastic} and \codebf{Rubber} are manually specified estimates.
Material presets are assigned according to asset semantics.
Building structures are modeled as \codebf{Concrete}; racks, shelf frames, pedestals, conveyors, and trolleys as \codebf{Metal}; pallets, cardboard boxes, crates, and goods as \codebf{Wood}; plastic barrels and containers as \codebf{Plastic}; and tires as \codebf{Rubber}.
Tab.~\ref{tab:acoustic_materials} summarizes the acoustic coefficients used for each material.
Across the 24 environments, this procedure assigns acoustic materials to 66240 mesh instances, consisting of 50851 \codebf{Wood}, 9465 \codebf{Metal}, 4943 \codebf{Concrete}, 636 \codebf{Plastic}, and 345 \codebf{Rubber} instances.

\begin{table}[t]
\centering
\caption{Acoustic material parameters used for geometry-dependent sound propagation. Abs.-L/M/H denote low-, mid-, and high-frequency absorption coefficients; Scat. denotes the scattering coefficient; and Trans.-L/M/H denote low-, mid-, and high-frequency transmission coefficients.}
\label{tab:acoustic_materials}
\resizebox{\columnwidth}{!}{%
\begin{tabular}{lccccccc}
\toprule
\toprule
Material & Abs.-L & Abs.-M & Abs.-H & Scat. & Trans.-L & Trans.-M & Trans.-H \\
\midrule
\rowcolor{uciblue!5}
Concrete & 0.05 & 0.07 & 0.08 & 0.05 & 0.015 & 0.002 & 0.001 \\
\rowcolor{uciblue!10}
Metal    & 0.20 & 0.07 & 0.06 & 0.05 & 0.200 & 0.025 & 0.010 \\
\rowcolor{uciblue!5}
Wood     & 0.11 & 0.07 & 0.06 & 0.05 & 0.070 & 0.014 & 0.005 \\
\rowcolor{uciblue!10}
Plastic  & 0.03 & 0.04 & 0.05 & 0.05 & 0.080 & 0.030 & 0.015 \\
\rowcolor{uciblue!5}
Rubber   & 0.05 & 0.10 & 0.15 & 0.10 & 0.020 & 0.010 & 0.005 \\
\bottomrule
\bottomrule
\end{tabular}%
}
\end{table}

At scene initialization, all eligible static meshes are transformed into world coordinates and merged into a single static acoustic representation, with one material assignment per object.
The acoustic mesh is constructed once per scene.
Skinned characters, skyboxes, and meshes attached to animated vehicles are excluded from this representation.
In particular, excluding animated vehicles prevents moving geometry from being incorrectly represented as stationary acoustic occluders at their initial poses.

\paragraph{Diffuse Industrial Noise Construction}
Industrial environments contain persistent acoustic interference from machinery, fans, ventilation systems, and other equipment.
We construct a spatially diffuse noise field using mono recordings from the MIMII dataset~\cite{purohit2019mimii}.

Recordings belonging to the same machine type are concatenated into a long waveform, with a $50$\,ms cross-fade applied at segment boundaries to reduce discontinuities.
For each episode, we sample $K=12$ non-overlapping waveform segments without replacement,
\begin{equation}
    \{n_i(t)\}_{i=1}^{K},
    \qquad K=12,
\end{equation}
providing temporally decorrelated noise signals for different spatial directions.

To approximate an isotropic diffuse sound field, we associate these signals with the $12$ vertices of a regular icosahedron:
\begin{equation}
    \mathcal{D}
    =
    \{\mathbf{d}_i\}_{i=1}^{12},
    \qquad
    \mathbf{d}_i \in \mathbb{S}^{2}.
\end{equation}
These directions provide approximately uniform coverage of the unit sphere.
At the beginning of each episode, we apply a random three-dimensional rotation
$\mathbf{R}\in SO(3)$,
\begin{equation}
    \widetilde{\mathbf{d}}_i
    =
    \mathbf{R}\mathbf{d}_i,
\end{equation}
which prevents persistent preferred noise directions across episodes.

\paragraph{FOA Encoding of the Diffuse Field}
Each noise component is interpreted as an independent plane wave.
Let
\begin{equation}
    \mathbf{q}_{i,t}
    =
    \left[
        q^{\mathrm{fwd}}_{i,t},
        q^{\mathrm{left}}_{i,t},
        q^{\mathrm{up}}_{i,t}
    \right]^\top
\end{equation}
denote the direction of the $i$-th component expressed in the current agent-centric coordinate system.
Using N3D normalization, the diffuse field is encoded as
\begin{align}
    W_t &= \sum_{i=1}^{K} n_i(t), \\
    X_t &= \sqrt{3}\sum_{i=1}^{K}
        q^{\mathrm{fwd}}_{i,t}\,n_i(t), \\
    Y_t &= \sqrt{3}\sum_{i=1}^{K}
        q^{\mathrm{left}}_{i,t}\,n_i(t), \\
    Z_t &= \sqrt{3}\sum_{i=1}^{K}
        q^{\mathrm{up}}_{i,t}\,n_i(t).
    \label{eq:diffuse_foa}
\end{align}
The resulting signal is stored using the ACN/N3D convention:
\begin{equation}
    \mathbf{X}^{\mathrm{noise}}_t
    =
    [W_t,Y_t,Z_t,X_t].
\end{equation}

Importantly, the diffuse field is anchored in the world coordinate system.
The plane-wave directions therefore remain fixed throughout an episode.
When the agent rotates, we transform the world-frame directions into its current egocentric coordinate system before recomputing Eq.~\ref{eq:diffuse_foa}.
A stationary agent therefore observes a spatially stable background field, whereas rotating the agent causes the field to counter-rotate in its local acoustic coordinates.
This avoids the physically implausible behavior in which the environmental noise field rotates together with the listener.
\paragraph{Noise-Level Calibration}
We parameterize the background-noise level using a nominal
signal-to-noise ratio (SNR). All level measurements are performed in the
target-signal frequency band, $\mathcal{B}=[500,900]$ Hz. Specifically,
for a waveform $x$ containing $T$ samples, we define the band-limited RMS
as
\begin{equation}
\operatorname{RMS}_{\mathcal{B}}(x)
=
\sqrt{
    \frac{1}{T}
    \sum_{t=1}^{T}
    \left[\mathcal{H}_{\mathcal{B}}(x)\right]^2_t
},
\label{eq:band_rms}
\end{equation}
where $\mathcal{H}_{\mathcal{B}}(\cdot)$ denotes a fourth-order
Butterworth band-pass filter over $\mathcal{B}$. Restricting the
measurement to this band prevents out-of-band energy in the machine
recordings from affecting the intended interference level.

\paragraph{Dataset-Level Reference Signal}
The nominal SNR is defined relative to a dataset-level reference signal
level, denoted by $\mathrm{REF}_{\mathrm{RMS}}$. We estimate this value
from the clean FOA observations of the corresponding acoustic condition.
For every scene, we randomly sample up to 60 trajectories and then sample
six observation steps from each selected trajectory. Importantly, the
steps are sampled across the trajectory rather than taken only from its
beginning, since the received target level generally increases as the
agent approaches the source. For each sampled observation, we compute the
band-limited RMS of the omnidirectional FOA channel $W$. The reference
level is the median over all sampled observations:
\begin{equation}
\mathrm{REF}_{\mathrm{RMS}}
=
\operatorname*{median}_{i\in\mathcal{S}}
\operatorname{RMS}_{\mathcal{B}}\!\left(
X^{\mathrm{clean}}_{i,W}
\right),
\label{eq:reference_rms}
\end{equation}
where $\mathcal{S}$ is the set of sampled observations and
$X^{\mathrm{clean}}_{i,W}$ is the clean $W$ channel of observation $i$.
We use the median because observations near the source can have
substantially higher energy and would disproportionately affect a mean
estimate. The reference level is computed separately for each acoustic
dataset and is shared by training and evaluation, ensuring that a given
nominal SNR corresponds to the same absolute noise level in both stages.

\paragraph{Noise-Field Normalization}
The sampled MIMII recordings are divided into non-overlapping streams,
and their episode-level band-limited RMS values are normalized before
spatial encoding. The normalized streams are assigned to the sampled
directions and jointly encoded into FOA. Because summing multiple
decorrelated streams can introduce small segment-to-segment level
variations, calibration is applied to the omnidirectional channel of the
aggregate FOA noise field. For a specified nominal SNR, the desired
background level is
\begin{equation}
\mathrm{RMS}_{\mathrm{noise}}
=
\frac{\mathrm{REF}_{\mathrm{RMS}}}
     {10^{\mathrm{SNR}_{\mathrm{nom}}/20}}.
\label{eq:noise_snr}
\end{equation}
The denominator uses $20$ rather than $10$ because the calibration is
expressed as an amplitude ratio between RMS values. Equivalently, the
corresponding power ratio is $10^{\mathrm{SNR}_{\mathrm{nom}}/10}$.

Let $\widetilde{\mathbf{X}}^{\mathrm{bg}}_t$ denote the FOA background
field constructed for observation segment $t$ before level calibration,
and let $\widetilde{X}^{\mathrm{bg}}_{t,W}$ denote its $W$ channel. The
segment-level calibration gain is
\begin{equation}
g_t
=
\frac{\mathrm{REF}_{\mathrm{RMS}}}
     {10^{\mathrm{SNR}_{\mathrm{nom}}/20}
      \operatorname{RMS}_{\mathcal{B}}
      \left(\widetilde{X}^{\mathrm{bg}}_{t,W}\right)},
\label{eq:noise_gain}
\end{equation}
and the calibrated background field is
\begin{equation}
\mathbf{X}^{\mathrm{bg}}_t
=
g_t\widetilde{\mathbf{X}}^{\mathrm{bg}}_t.
\end{equation}
The same gain is applied jointly to all four FOA channels, preserving
their spatial relationships.

\paragraph{Nominal and Realized SNR}
The nominal SNR controls the absolute background-noise floor; it does not
force the target-to-noise ratio of every observation to equal the
specified value. In closed-loop episodes, one nominal SNR is sampled and held fixed for the entire episode. The realized SNR at observation
segment $t$ is
\begin{equation}
\mathrm{SNR}^{\mathrm{real}}_t
=
20\log_{10}
\frac{
\operatorname{RMS}_{\mathcal{B}}
\left(X^{\mathrm{clean}}_{t,W}\right)
}{
\operatorname{RMS}_{\mathcal{B}}
\left(X^{\mathrm{bg}}_{t,W}\right)
}.
\label{eq:realized_snr}
\end{equation}
Although the background field is normalized segment by segment to
maintain the prescribed absolute noise floor, this normalization depends
only on the sampled noise and never on the instantaneous target level.
It is therefore not an automatic gain-control procedure that enforces a
constant target-to-noise ratio. As the robot approaches the target, the
received target signal generally becomes stronger while the calibrated
noise floor remains unchanged, causing
$\mathrm{SNR}^{\mathrm{real}}_t$ to increase naturally. This preserves
the physical relationship between source distance and acoustic
observability.

\begin{table*}[t]
\centering
\caption{Acoustic characteristics of the target and localized equipment-noise sources.
SPL$_\mathrm{snd}$ values are nominal levels at 1\,m under our waveform calibration.
The 10--90\% energy band denotes the frequency range containing the central 80\% of spectral energy, and the overlap column reports the fraction of total energy within the target chirp band (500--900\,Hz).}
\label{tab:localized_noise}
\resizebox{.95\textwidth}{!}{%
\begin{tabular}{lccccc}
\toprule
\toprule
Source
& \shortstack{SPL$_\mathrm{snd}$@1\,m\\unweighted}
& \shortstack{SPL$_\mathrm{snd}$@1\,m\\A-weighted}
& \shortstack{10--90\%\\Energy Band}
& \shortstack{Energy in\\500--900\,Hz}
& \shortstack{Temporal\\Structure} \\
\midrule
Target      & 95.1 & 93.6 & 560--840\,Hz   & 96.0\% & 50\,ms sweep, 20\,Hz \\
Conveyor    & 72.0 & 65.6 & 178--477\,Hz   & 5.8\%  & steady hum \\
Forklift    & 68.0 & 58.4 & 110--265\,Hz   & 0.15\% & steady hum \\
Reach truck & 66.0 & 60.6 & 183--443\,Hz   & 5.6\%  & steady hum \\
AGV         & 62.0 & 60.7 & 551--1480\,Hz  & 74.7\% & steady hum \\
\bottomrule
\bottomrule
\end{tabular}%
}
\end{table*}

\paragraph{Localized Noise Construction}
We augment the industrial scenes with localized noise emitters attached to machinery and mobile platforms.
The audio recordings are sourced from~\citet{lumamorph_engine_idling_2022}.
Across the 24 scenes, we instantiate 190 localized noise sources: 81 automated guided vehicles (AGVs), 57 conveyors, 31 forklifts, and 21 reach trucks, corresponding to 4--18 sources per scene (7.9 on average).
Among these sources, 117 are attached to objects that move during an episode, while the remaining 73 correspond to static machinery or parked vehicles.
Source poses are updated at every simulation step, and all localized noise sources are rendered through the same acoustic propagation pipeline as the navigation target, including reflection, occlusion, and transmission effects.
Thus, their received spatial signatures vary jointly with source motion, agent pose, and scene geometry.

We use one representative mono audio loop for each equipment category.
All clips are stored at 48\,kHz and are processed into seamless loops whose lengths are integer multiples of the 1,024-sample audio frame.
The processing pipeline consists of high-pass filtering for DC removal, resampling or pitch adjustment when needed, temporal cropping, frame-aligned trimming, equal-power crossfading at the loop boundary, and final amplitude normalization.
The resulting loop durations are 2.411\,s, 8.341\,s, 4.992\,s, and 3.264\,s for the conveyor, forklift, reach truck, and AGV, respectively.

The source amplitudes are fixed directly in the audio waveforms, with no additional runtime gain.
For simulation calibration, we define 0\,dBFS to correspond to a nominal sound pressure level (SPL$_\mathrm{snd}$\footnote{We use SPL$_\mathrm{snd}$ to denote sound pressure level, distinguishing it from the navigation metric SPL (Success weighted by Path Length).}) of 100\,dB at 1\,m.
Under this convention, the conveyor, forklift, reach truck, and AGV have nominal unweighted levels of 72, 68, 66, and 62\,dB SPL$_\mathrm{snd}$ at 1\,m, respectively, while the target chirp has a nominal level of 95.1\,dB SPL$_\mathrm{snd}$ at 1\,m.
The equipment-noise levels are chosen to be broadly consistent with representative real-world industrial machinery, while the target is intentionally configured as a salient industrial acoustic signal~\cite{yale_erp13_15vc, mitsubishi_sensia_em, mir100_specs, bosch_varioflow_belt, rollco_timing_belt_conveyor, iso7731_2003}.

Tab.~\ref{tab:localized_noise} summarizes the level, spectral, and temporal characteristics of the target and localized noise sources.
\textit{SPL$_\mathrm{snd}$@1\,m} denotes the nominal unweighted SPL$_\mathrm{snd}$ derived from the waveform RMS under the above calibration convention.
\textit{A-weighted SPL$_\mathrm{snd}$} applies standard perceptual frequency weighting and is reported only to facilitate comparison with equipment specifications commonly expressed in dB(A).
The \emph{10--90\% energy band} denotes the frequency interval containing the central 80\% of signal energy, while \emph{energy in 500--900\,Hz} measures the fraction of total spectral energy that overlaps the target chirp band.
The final column summarizes the dominant temporal structure of each source.

The unweighted SPL$_\mathrm{snd}$ is the level used by the simulator and is therefore the quantity directly experienced by the navigation agent and used in defining the diffuse-noise SNR settings.
A-weighted SPL$_\mathrm{snd}$ is reported only for comparison with real equipment specifications.
Because A-weighting attenuates low-frequency energy, it reduces the apparent level of the conveyor, forklift, and reach-truck hums more strongly than that of the higher-frequency target and AGV signals.
The target is intentionally designed to remain acoustically salient relative to localized machinery noise.
Its 93.6\,dB(A) nominal level is substantially above the loudest localized source, while its periodic chirp structure also differs from the predominantly steady equipment hums.
This design follows the general principle of industrial auditory signaling, where warning or guidance signals should remain distinguishable from background machinery in level, spectral content, and temporal structure.

\paragraph{Episode-wise SNR Sampling}
For the \codebf{DN} condition, the nominal background-noise SNR is sampled independently for each episode and then held fixed throughout that episode. 
Thus, the \codebf{DN} results reported in the main navigation tables do not correspond to a single fixed SNR, but instead reflect performance under an episode-wise distribution of background-noise levels.
The \codebf{LN} condition follows the same SNR sampling for its diffuse background component and further adds the localized object-associated noise sources described below.

\paragraph{Expert Trajectory Collection} Following the protocol in~\cite{li2025industrynav}, expert trajectories were collected using an A* expert.
Start–target pairs were sampled from occupancy maps and retained only when both endpoints had sufficient obstacle clearance and a feasible path existed. 
At each state, the expert used the current top-down viewed occupancy observation to select a discrete navigation action, while the collector recorded the corresponding depth image, first-order ambisonic audio, and action label. An episode ended with a recorded stop action when the agent came within the neighborhood of the target sound source.
\section{Reinforcement Learning Details}
\label{app:method}
\subsection{Backbone-Shared Actor--Critic}
\label{app:actor-critic}
PPO requires both a policy for action selection and a value function for estimating future returns. A straightforward implementation with separate multimodal actor and critic networks, however, would duplicate the computationally expensive audio--depth backbone. Conversely, fully sharing the backbone while allowing both policy and value losses to update it can introduce gradient interference, particularly when the critic is still adapting to the return scale. We therefore adopt a \emph{backbone-shared but gradient-decoupled} actor--critic architecture, where actor and critic reuse the same multimodal representation during the forward pass while maintaining separate optimization paths.

Given the current audio--depth observation and navigation prompt, the shared backbone produces a contextual hidden representation $\mathbf{h}_{t}$ at the final prompt position, immediately before the LMM output head. This position is used by both branches because it has attended to the complete multimodal prompt and is also the representation from which the LMM predicts the navigation response. The actor treats the LMM as a discrete navigation policy rather than performing unrestricted text generation. Specifically, we retain only the logits associated with the valid navigation answer tokens, e.g., \texttt{forward}, \texttt{left}, \texttt{right}, and \texttt{stop}, and construct a categorical distribution over these actions. To enable lightweight policy adaptation during PPO, we augment the original SFT action logits with a trainable residual action head:
\begin{align}
&\mathbf{z}_t
=
\mathbf{z}^{\mathrm{SFT}}_t
+
\Delta W\mathbf{h}_t, \notag
\\
&\pi_\theta(a_t\mid O_t, a_{\leq t-1})
=
\operatorname{Softmax}(\mathbf{z}_t), \notag
\end{align}
where $\mathbf{z}^{\mathrm{SFT}}_t$ contains the original language-model logits corresponding only to the valid action tokens, and $\Delta W$ directly adjusts their relative preferences. The residual head is initialized to zero, such that $\mathbf{z}_t=\mathbf{z}^{\mathrm{SFT}}_t$ before RL updates and PPO therefore starts exactly from the SFT policy rather than from a randomly initialized action classifier. 

The critic operates on the same final-token representation $\mathbf{h}_t$, but predicts a scalar state value through a lightweight value head. Specifically, the critic receives $\operatorname{stopgrad}(\mathbf{h}_t)$, applies layer normalization followed by an MLP, and predicts the scalar state value. 
Since our navigation task has a finite episode horizon, we additionally make the critic aware of the remaining action budget,
\[
b_t=\frac{T_{\max}-t}{T_{\max}},
\]
where $t$ denotes the number of actions already executed. 
The value function is therefore parameterized as
\begin{equation}
    V_\psi(s_t,b_t)
    =
    f_\psi\!\left(
    \operatorname{stopgrad}(\mathbf{h}_t),
    b_t
    \right).\notag
\end{equation}
This temporal signal is necessary because the \textbf{same audio--depth observation can correspond to different expected returns depending on how much navigation budget remains}. 
For example, otherwise identical observations encountered at the beginning and the end of navigation provide the critic with nearly the same multimodal representation but have substantially different remaining horizons. 
Without $b_t$, the critic would be forced to assign a single value to these two situations, making its input representation insufficient for estimating the finite-horizon return. 
The remaining-budget feature is therefore \textbf{provided exclusively to the critic}, while the actor remains time-agnostic and receives exactly the same multimodal inputs as the original SFT policy. 

Importantly, backbone sharing is restricted to representation reuse rather than unrestricted gradient sharing. We detach $\mathbf{h}_t$ before passing it to the value head, so gradients from the value loss update only the critic and do not propagate through the shared backbone. The actor, in contrast, continues to optimize the navigation policy through the backbone and residual action head. This separation prevents the critic objective from dominating or distorting policy-relevant representations while retaining the computational advantage of evaluating the multimodal backbone only once for both branches. 

\subsection{Action-Space Potential Construction}
\label{app:action_potential}

TARS requires the minimum remaining number of discrete actions
$N(s)$ from a simulator state $s$ to the goal region.
We precompute this quantity from the scene traversability map in the
same discrete action space used by the navigation policy.

Specifically, we first inflate obstacles in the scene occupancy map
and downsample the resulting free-space representation into a navigation
grid. Each spatial grid location is augmented with a discrete orientation,
forming a state graph over position--heading pairs. Since the policy rotates
in increments of $18^\circ$, the heading space contains 20 orientation bins,
aligned with the initial heading of the corresponding episode.

Each graph edge corresponds to one navigation action:
\texttt{forward}, \texttt{left}, or \texttt{right}, with unit action cost.
The rotation actions transition between adjacent heading bins, whereas
\texttt{forward} advances the agent by approximately $1.5\,\mathrm{m}$
along its current heading. A forward edge is included only when the
corresponding motion path is collision-free.

To obtain the action distance, all traversable position--heading states
whose positions lie within the goal radius $\delta$ are initialized with
$N(s)=0$, irrespective of orientation. We then perform a reverse
breadth-first search over the state graph. Since every edge has unit cost,
the resulting value $N(s)$ gives the minimum number of discrete navigation
actions required to reach the goal region from each reachable
position--heading state.

During interaction, the simulator state is continuous and therefore does
not necessarily coincide exactly with a grid state. For a query state, we
use the minimum $N(s)$ among traversable grid locations within a radius of
two grid cells having the same discretized heading. If no valid state is
available in this neighborhood, we fall back to the nearest traversable
grid location. This lookup is used only to construct the training reward;
the resulting action distance and the underlying occupancy map are never
provided to the policy.

\begin{figure*}[t]
\centering
\resizebox{\textwidth}{!}{
\includegraphics{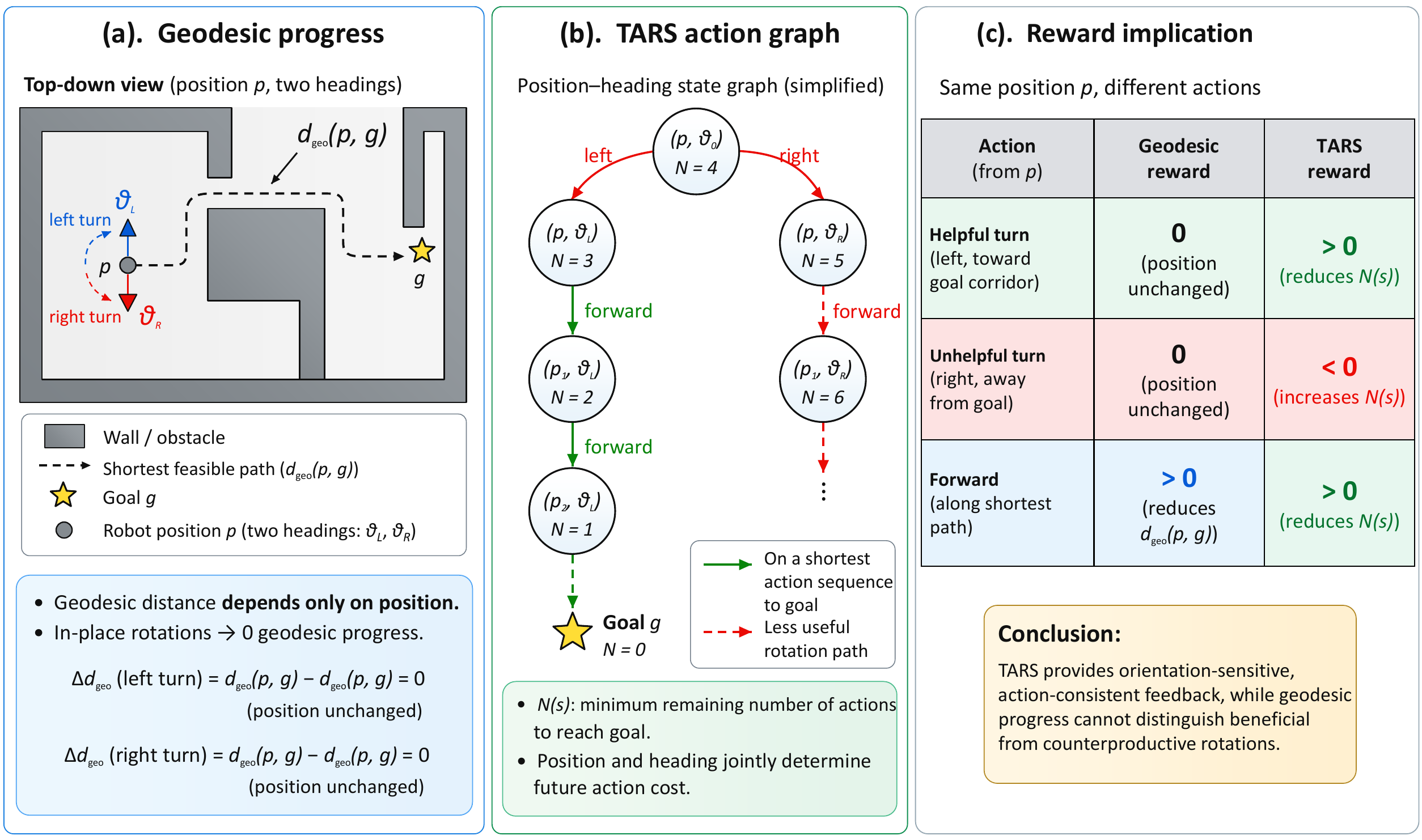}
}
\caption{\textbf{Comparison of geodesic progress and TARS.} (a). In-place rotations leave geodesic distance unchanged, providing no feedback on heading changes. (b). TARS uses a position–heading action graph to measure the minimum remaining action count ($N(s)$). (c). With $\gamma_\Phi=1$, TARS rewards turns that reduce $N(s)$ and penalizes those that increase it, distinguishing helpful from counterproductive rotations at the same position.}
\label{fig:tars}
\end{figure*}

As illustrated in Fig.~\ref{fig:tars}(a),
in-place rotations yield zero geodesic progress despite
their different effects on future navigation effort. 
TARS captures these differences through a position--heading action graph, providing
orientation-sensitive feedback based on changes in the minimum remaining action count
(Fig.~\ref{fig:tars}(b,c)).

\subsection{PPO Optimization}
\label{app:ppo}
We optimize the navigation policy using clipped PPO.
Given trajectories collected under the behavior policy
$\pi_{\theta_{\mathrm{old}}}$,
we estimate advantages using generalized advantage estimation (GAE).
For a non-terminal transition,
\begin{equation}
    \delta_t
    =
    r_t
    +
    \gamma
    V_\psi(s_{t+1},b_{t+1})
    -
    V_\psi(s_t,b_t),
\end{equation}
with bootstrapping masked at terminal transitions.
The advantage estimate is
\begin{equation}
    \hat{A}_t
    =
    \sum_{l=0}^{T-t-1}
    (\gamma\lambda)^l
    \delta_{t+l},
\end{equation}
where $T$ denotes the terminal step of the collected trajectory.

The PPO probability ratio is
\begin{equation}
    \rho_t
    =
    \frac{
        \pi_\theta(a_t\mid s_t)
    }{
        \pi_{\theta_{\mathrm{old}}}(a_t\mid s_t)
    },
\end{equation}
and the clipped policy objective is
\begin{equation}
\mathcal{L}_{\pi}
=
-
\mathbb{E}_t
\left[
\min
\left(
\rho_t\hat{A}_t,\,
\mathrm{clip}
(\rho_t,1-\epsilon,1+\epsilon)
\hat{A}_t
\right)
\right].
\end{equation}

The critic is optimized with a clipped value objective.
Let
\begin{align}
    &V_\psi^{\mathrm{clip}}(s_t,b_t)
    =
    V_{\psi_{\mathrm{old}}}(s_t,b_t)
    + \notag \\
    &\mathrm{clip}
    \left(
        V_\psi(s_t,b_t)
        -
        V_{\psi_{\mathrm{old}}}(s_t,b_t),
        -\epsilon_V,
        \epsilon_V
    \right).
\end{align}
Then,
\begin{align}
\mathcal{L}_{V}
=
\frac{1}{2}
\mathbb{E}_t
\Big[
\max
\Big(
&
\big(
V_\psi(s_t,b_t)-\hat{R}_t
\big)^2,
\notag\\
&
\big(
V_\psi^{\mathrm{clip}}(s_t,b_t)-\hat{R}_t
\big)^2
\Big)
\Big],
\end{align}
where $\hat{R}_t$ denotes the return target.
An entropy bonus is included to encourage exploration, yielding
\begin{equation}
\mathcal{L}_{\mathrm{PPO}}
=
\mathcal{L}_{\pi}
+
c_V\mathcal{L}_{V}
-
c_H\mathcal{H}(\pi_\theta).
\end{equation}

\paragraph{SFT Anchoring}
Although PPO improves closed-loop behavior on states induced by the learned policy itself, unconstrained RL updates may cause the policy to drift away from the navigation behavior acquired during SFT.
We therefore retain a frozen copy of the SFT policy,
$\pi_{\mathrm{SFT}}$, as a reference policy.
For states encountered during RL rollouts, we regularize the current navigation distribution using
\begin{equation}
\mathcal{L}_{\mathrm{KL}}
=
\mathbb{E}_{s_t\sim\pi_\theta}
\left[
D_{\mathrm{KL}}
\left(
\pi_{\mathrm{SFT}}(\cdot\mid s_t)
\;\|\;
\pi_\theta(\cdot\mid s_t)
\right)
\right].
\end{equation}
Both distributions are defined over the same restricted navigation action set rather than the full language-model vocabulary.
The final reinforcement-learning objective is
\begin{equation}
\mathcal{L}_{\mathrm{RL}}
=
\mathcal{L}_{\mathrm{PPO}}
+
\beta\mathcal{L}_{\mathrm{KL}},
\label{eq:loss_rl}
\end{equation}
where $\beta$ controls the strength of the SFT constraint.
Together with the zero initialization of the residual action head, this regularization anchors PPO to the SFT policy while still allowing the action preferences and policy representation to adapt to closed-loop navigation experience.

\paragraph{Episode Termination}
Episodes terminate when the agent executes \texttt{stop} or reaches the maximum step budget $T_{\max}=100$.
A navigation episode is considered successful only when \texttt{stop} is issued within the target radius:
\begin{equation}
\mathrm{succ}
=
\mathbbm{1}
\{a_t=\texttt{stop}\}
\mathbbm{1}
\{d_{\mathrm{euc}}(s_t,g)\le r_{\mathrm{goal}}\}.
\end{equation}
Issuing \texttt{stop} outside the target region terminates the episode without the terminal success reward.
Unless otherwise specified in an ablation, we do not require an additional wrong-stop penalty: premature stopping is already discouraged because it forfeits subsequent transition-shaping rewards as well as the terminal success reward.

\paragraph{Reward Configuration}
For PPO training, we use a per-step slack penalty of
$r_{\mathrm{slack}}=-0.01$ and a terminal success scale of
$R_{\mathrm{succ}}=10.0$. Consequently, a successful episode receives the terminal reward
\[
r_{\mathrm{succ}} = 10.0 \times \mathrm{SPL}.
\]
For the transition-aware potential term in Eq.~\eqref{eq:tars},
we set $\gamma_{\Phi}=1$.

\section{Evaluation Metrics}
\label{app:metrics}
We evaluate navigation performance using Success Rate (SR), Success weighted by Path Length (SPL), Success weighted by Number of Actions (SNA), and Final Distance to Goal (FDG), following standard evaluation protocols in embodied and audio-guided navigation.
Together, these metrics evaluate three complementary aspects of navigation: task completion, geometric path efficiency, and discrete action efficiency.

Let $N$ denote the total number of evaluation episodes.
For the $i$-th episode, we define a binary success indicator
\begin{equation}
    S_i =
    \mathbbm{1}\{a_i^{\mathrm{final}}=\texttt{stop}\}\mathbbm{1}
    \{
        d_i^{\mathrm{final}} \leq \delta
    \},
    \label{eq:success_indicator}
\end{equation}
where $d_i^{\mathrm{final}}$ is the final distance between the agent and the acoustic target, $a_i^{\mathrm{final}}$ is the final action, and $\delta$ is the predefined success radius.
Thus, $S_i=1$ indicates a successful episode and $S_i=0$ otherwise.

\paragraph{Success Rate}
SR directly measures the agent's ability to reach the acoustic
target:
\begin{equation}
    \mathrm{SR}
    =
    \frac{1}{N}
    \sum_{i=1}^{N} S_i.
    \label{eq:sr}
\end{equation}
SR measures task completion independently of how efficiently the target is reached.
An agent that takes a highly circuitous trajectory therefore receives the same episode-level success score as an agent that follows an optimal route, provided that both eventually reach the target.
While SR provides the most direct measure of navigation reliability, it does not capture the efficiency of successful trajectories.

\paragraph{Success weighted by Path Length}
To jointly account for task success and geometric navigation efficiency, we adopt SPL.
Let $L_i$ denote the length of the shortest feasible path from the initial agent position to the target in episode $i$, and let $P_i$ denote the length of the trajectory actually traveled by the agent.
SPL is defined as
\begin{equation}
    \mathrm{SPL}
    =
    \frac{1}{N}
    \sum_{i=1}^{N}
    S_i
    \frac{L_i}
    {\max(P_i,L_i)}.
    \label{eq:spl}
\end{equation}
Failed episodes contribute zero regardless of their trajectory length.
For a successful episode, the score approaches one when the executed trajectory follows the shortest feasible route and decreases as the agent takes increasingly circuitous paths.
The $\max(P_i,L_i)$ term bounds the per-episode score by one and makes the
metric robust to small numerical discrepancies between computed and
executed path lengths.
Importantly, $L_i$ corresponds to the shortest \emph{navigable} path rather
than the straight-line Euclidean distance between the start and target.
We compute this quantity using the simulator's traversability information so that obstacles and scene geometry are taken into account.

\paragraph{Success weighted by Number of Actions}
SPL alone does not fully characterize efficiency in a discrete-action navigation setting.
For example, an agent may repeatedly rotate in place or attempt forward motions that result in collisions.
Such actions increase computational and interaction cost but may contribute little or no additional geometric path length.
We therefore additionally report SNA.

Let $A_i^{*}$ denote the minimum number of discrete actions required to successfully reach the target from the initial state of episode $i$, under the same action space and transition rules used by the evaluated agent.
Let $A_i$ denote the total number of actions actually executed by the agent.
SNA is defined as
\begin{equation}
    \mathrm{SNA}
    =
    \frac{1}{N}
    \sum_{i=1}^{N}
    S_i
    \frac{A_i^{*}}
    {\max(A_i,A_i^{*})}.
    \label{eq:sna}
\end{equation}
As with SPL, failed episodes receive a score of zero.
A successful episode achieves an SNA score of one when the agent reaches the target using the minimum possible number of actions.
The score decreases as additional actions are introduced.

Unlike SPL, SNA explicitly accounts for actions that do not necessarily change the agent's spatial position.
For instance, consider two agents that traverse nearly identical geometric paths.
If one agent performs several unnecessary left--right rotations before each forward motion, the two trajectories can have similar SPL values, while the second agent receives a lower SNA score.
Similarly, unsuccessful forward attempts caused by collisions increase the executed action count even if the agent remains at the same location.
SNA therefore complements SPL by capturing inefficiencies arising from redundant decision-making and poor low-level control.

\paragraph{Final Distance to Goal} FDG measures the Euclidean distance between the agent's position and the target position at episode termination, either when the agent issues \texttt{stop} or when the maximum step budget is reached.
FDG is defined as
\begin{equation}
    \text{FDG} = \frac{1}{M}\sum_{i=1}^{M}||\mathbf{p}^{s}_i - \mathbf{p}^{end}_i||_2,
\end{equation}
where \(M\) is the number of episodes, \(\mathbf{p}^{end}_{i}\) is the agent's position when the $i-$th episode ends, \(\mathbf{p}^{s}_{i}\) is the corresponding target position.

\paragraph{Metric Complementarity.}
The four metrics emphasize distinct properties of the navigation policy.
SR evaluates whether the target is successfully reached, while FDG measures how close the agent is to the target at episode termination, providing a continuous measure of navigation progress even for unsuccessful episodes.
SPL evaluates whether successful navigation follows an efficient geometric route, and SNA evaluates whether that route is realized through an efficient sequence of discrete actions.
Consequently, a strong navigation policy should achieve high SR, SPL, and SNA while maintaining a low FDG, rather than optimizing any single measure in isolation.

\section{Additional Implementation Details}
\label{app:impl_details}
\subsection{Spatial-Audio Adaptation}
\label{app:training_saa}
In the first stage, we adapt the pretrained audio representation to spatial audio using the proposed SAA together with an auxiliary direction-prediction head.
The pretrained Qwen2.5-Omni audio encoder, vision encoder, multimodal projectors, and base language-model parameters remain frozen during this stage.
Only the SAA and the auxiliary direction head are optimized.
The auxiliary head is used solely to provide explicit directional supervision during spatial-audio adaptation and is discarded after Stage~1, while the trained SAA is retained for subsequent multimodal navigation training.
We use AdamW with a learning rate of
$1\times10^{-4}$,
weight decay $0.01$,
and betas $(0.9,0.999)$.
The learning rate follows cosine decay with a $5\%$ warmup period.
We use gradient clipping at a maximum norm of $1.0$.
The per-device batch size is $8$,
with $8$ gradient accumulation steps,
resulting in an effective batch size of $64$.
We train using bfloat16 precision on a single GPU.

\subsection{Multimodal Supervised Fine-Tuning}
\label{app:sft_training}
We perform SFT for navigation action prediction using spatial audio and egocentric depth observations.
The SAA learned in Stage~1 is retained and further optimized, while the pretrained audio tower, vision tower, native multimodal projectors, and Qwen2.5-Omni base weights remain frozen.
We additionally introduce LoRA adapters~\cite{hu2021lora} into the LMM decoder.
LoRA is applied only to the self-attention projections
\texttt{q\_proj}, \texttt{k\_proj}, \texttt{v\_proj}, and \texttt{o\_proj},
with rank $r=8$, scaling factor $\alpha=16$, and dropout $0.05$.
We use AdamW with learning rate
$1\times10^{-4}$,
weight decay $0.01$,
and betas $(0.9,0.999)$.
The learning rate follows cosine decay with a $5\%$ warmup period.
Gradients are clipped to a maximum norm of $1.0$, and the effective batch size is $64$.
The per-device batch size is $2$,
with $32$ gradient accumulation steps
on a single GPU.

\subsection{PPO Fine-Tuning}
\label{app:ppo_training}
Starting from the SFT checkpoint, we further optimize the navigation policy with PPO through closed-loop interaction with the Unity environment. Unlike SFT, which learns from one-step expert demonstrations, PPO exposes the policy to states induced by its own decisions and optimizes long-horizon navigation performance under the reward defined in Sec.~\ref{subsec:rl}. We use the same discrete navigation action interface during RL and inference.

\paragraph{Trainable Parameters}
During PPO fine-tuning, the SAA, audio tower, vision tower, and Qwen base model remain frozen. The trainable actor parameters consist of the LMM LoRA parameters and an action-logit residual
$\Delta W$.
As described in App.~\ref{app:actor-critic}, $\Delta W$ directly adjusts the logits over the restricted navigation action set and is initialized to zero, ensuring that the initial RL policy exactly recovers the SFT policy.
The critic uses a separate scalar value head attached to the shared final-token representation. We implement the value head in FP32 as a LayerNorm--MLP and a SiLU activation. 

\paragraph{Optimization Hyperparameters}
We use a discount factor of $\gamma=0.99$ and GAE parameter
$\lambda=0.95$.
The PPO clipping coefficient is $0.2$, the entropy coefficient is $0.01$, and the value-loss coefficient is $0.5$.
Advantages are normalized within each PPO update, and we use the clipped value objective described in App.~\ref{app:ppo}.
Each collected rollout is optimized for one PPO epoch.

We use Adam with $\epsilon=10^{-5}$ and clip the gradient norm to $0.5$ independently for each parameter group.
Because the backbone adaptation, critic, and action residual have substantially different parameterizations and optimization dynamics, we assign separate learning rates:
$1\times10^{-5}$ for the LLM LoRA parameters,
$1\times10^{-3}$ for the value head, and
$1\times10^{-6}$ for the action-logit residual.

\paragraph{Rollout and Update Schedule}
Training uses fixed-length on-policy rollouts collected from parallel Unity environments.
Each rollout chunk contains $T=16$ environment steps from each of $N$ parallel environments, and we collect $K=2$ chunks for each PPO update.
Thus, each update contains
\begin{equation}
    N_{\mathrm{trans}}
    =
    T N K
\end{equation}
on-policy transitions.
The resulting rollout buffer is divided into 16 minibatches with a microbatch size of 10 for gradient accumulation.

Episodes need not align with rollout boundaries.
When an episode terminates inside a rollout, either by executing \texttt{stop} or reaching the maximum horizon, the corresponding terminal mask prevents value bootstrapping across the boundary and the environment is immediately reset.
Consequently, a rollout buffer may contain transitions from multiple consecutive episodes while remaining entirely on-policy.

\paragraph{Critic Warm-Up}
We calibrate the value function before allowing PPO to modify the actor.
During this warm-up stage, the actor is frozen and trajectories are generated by the unchanged SFT policy, while only the value head is optimized against returns induced by the RL reward.
This allows the critic to learn the scale and structure of the return signal before its estimates are used to drive policy updates.
We perform at most 100 value-only updates.
Warm-up terminates early once the critic achieves an explained variance greater than $0.5$ for three consecutive updates, after which training switches to joint PPO optimization.
This criterion avoids unnecessarily extending critic pretraining once the value function provides a sufficiently informative baseline while preventing poorly calibrated early advantages from producing destructive updates to the SFT policy.

We adapt the KL coefficient around a target divergence of $0.06$.
The coefficient in Eq.~\ref{eq:loss_rl} is initialized to $\beta=0.2$ and constrained to
$[10^{-4},10]$.
When the observed divergence exceeds the target, $\beta$ is increased by a factor of $2.0$; when it falls below the target, it is reduced by a factor of $1.1$.
We use a 50-update arming period before enabling this adaptive controller, allowing the early PPO dynamics to stabilize before adjusting the strength of the KL constraint.

\paragraph{Overall Training Procedure}
The RL stage therefore proceeds in two phases.
We first perform critic-only warm-up while retaining the original SFT actor.
After the value function reaches the explained-variance criterion, we repeatedly
(i) collect fixed-length on-policy trajectories from parallel environments,
(ii) compute bootstrapped returns and GAE advantages,
(iii) optimize the actor and critic with PPO minibatches.
This procedure retains the strong behavior learned from demonstrations while allowing the policy to adapt to the distribution of states created by its own closed-loop actions and to optimize long-horizon navigation quality directly.

\subsection{Action-History Context}
For configurations using action history, each prompt contains the eight
most recent navigation actions in chronological order. At the beginning
of an episode, unavailable entries are represented by \texttt{none}.
The history contains actions only; no previous audio, depth observation,
collision outcome, or privileged state is retained.

\section{Zero-Shot Multimodal Baselines}
\label{app:zero_shot_baselines}

We additionally evaluate general-purpose LMMs without any task-specific training on \dname{}.
These models receive no supervised demonstrations, RL, or navigation-specific parameter updates from our benchmark.
Instead, they directly map multimodal observations to the
discrete navigation action space used by \mname{}.
We evaluate two settings: a standard zero-shot without oracle information
setting and a zero-shot with assisted oracle information setting. The former measures
whether general-purpose multimodal capabilities transfer directly to
audio-depth navigation, while the latter isolates downstream action
reasoning from spatial-audio perception.

The oracle-assisted zero-shot model uses the full action space, and the observation-only model uses the motion-action subset.

\paragraph{Multimodal Input Interface.}
General-purpose LMMs do not natively accept an FOA observation as a single four-channel audio tensor.
We therefore expose the four FOA channels through the models'
standard multi-audio interface.
Specifically, the channels are provided as four separate mono audio
inputs in the fixed ACN/N3D order
\[
[W,Y,Z,X],
\]
where $W$ is the omnidirectional component, $X$ corresponds to the
front--back axis, $Y$ to the left--right axis, and $Z$ to the vertical
axis.
The prompt explicitly specifies this channel convention.
We neither downmix the FOA observation into mono nor convert it into
binaural audio.
The egocentric depth observation is provided as an image together
with the four audio inputs.

\paragraph{Zero-Shot without Oracle Information}
In the standard zero-shot setting, the model receives only the current
FOA observation, the current egocentric depth image, and a bounded
history containing the eight most recent actions commanded by the
agent.
No previous audio or depth observations are retained.
Moreover, the model is not provided with source bearing, source
distance, agent pose, source position, map coordinates, shortest-path
information, collision outcomes, or any other ground-truth navigation
state.

The action history is used only as a lightweight control prior.
It records previously issued commands rather than their outcomes and
therefore cannot directly reveal whether a movement succeeded,
whether the agent made progress, or whether an earlier directional
estimate was correct.
Its purpose is primarily to reduce short-horizon failure patterns such
as repeated turns or left--right oscillation while preserving the
partially observable nature of the task.

The prompt instructs the model to infer the target direction from the
current FOA observation, identify immediately traversable space from
the current depth image, and select one of
\texttt{forward}, \texttt{left}, or \texttt{right}.
We disable \texttt{stop} in this setting because the observation-only
model is not given a reliable metric estimate of source distance.
Instead, we adopt a relaxed stopping criterion
for this diagnostic setting. Specifically, the \texttt{stop} action is
disabled, and an episode is automatically counted as successful once the
agent enters the target region,
\[
d_t^{s} \leq \delta.
\]
Thus, observation-only zero-shot agents are evaluated on whether they can
navigate into the goal neighborhood, without additionally requiring them
to infer when to terminate from acoustic amplitude. This protocol is
intentionally favorable to the zero-shot baselines and isolates their
ability to recover target direction and perform obstacle-aware navigation.

\paragraph{Zero-Shot with Oracle Information}
We further evaluate an oracle-assisted condition in which the LMM
receives the ground-truth agent-centric bearing of the target and its
straight-line Euclidean distance from the agent, in addition to the
current depth and FOA observations.
The bearing follows the convention that $0^\circ$ denotes a target
directly ahead, positive values indicate the target is to the left,
and negative values indicate the target is to the right.
The oracle distance additionally enables the model to determine when
the agent is within the success radius and should issue
\texttt{stop}.

This condition is \emph{not} intended as a directly comparable
navigation baseline, because it exposes privileged information that
is unavailable to \mname{} during inference.
Instead, it serves as a diagnostic decomposition of zero-shot
navigation performance.
By removing the need to recover target direction and distance from
the acoustic observation, the oracle setting approximately isolates
the model's ability to combine goal direction with local depth
geometry and select appropriate navigation actions.
Consequently, the performance gap between the oracle-assisted and
observation-only settings provides evidence about whether the primary
zero-shot bottleneck arises from spatial perception or from downstream
action reasoning.

\paragraph{Closed-Loop Evaluation.}
All zero-shot models are evaluated through the same closed-loop
environment interaction protocol.
At each timestep, the model receives the currently available
multimodal inputs and predicts a single discrete action.
The selected action is executed in the Unity environment, after which
the simulator generates a new audio-depth observation and the model
is queried again.
No future observation, transition outcome, or privileged simulator
state is revealed before the next decision.
Evaluation uses the same navigation dynamics and maximum episode
horizon as the other methods.

For the observation-only setting, the \texttt{stop} action is disabled.
An episode automatically terminates successfully once the agent enters
the target radius $\delta$; otherwise, it terminates when the maximum
navigation horizon is reached. For the oracle-assisted setting, the model
may explicitly emit \texttt{stop}, and success follows the standard strict
stopping criterion used by \mname.

\paragraph{Common Prompt Design.}
To minimize prompt-dependent variation, all zero-shot models receive
the same task definition and modality descriptions.
The common instructions specify:
(i) the FOA channel order and spatial semantics;
(ii) the depth-image encoding and egocentric viewpoint;
(iii) the discrete navigation action semantics; and
(iv) the required one-action output format.
The observation-only prompt additionally states the information
boundary described above and explains that action history contains
commands rather than verified state transitions.
The oracle-assisted prompt instead identifies the provided bearing
and distance as authoritative navigation information.
Rather than encouraging unrestricted chain-of-thought generation,
the models are required to return only one valid navigation action.
This minimizes differences arising from output verbosity and enables
a common action interface across model families.

We use identical decoding and response-processing rules across
zero-shot LMMs. No model-specific navigation heuristic or manually designed
post-processing rule is applied.

\paragraph{Evaluated Models}
The zero-shot baselines included in the main experiments are
Qwen2.5-Omni and Qwen3-Omni.
Both models are evaluated under the same multimodal input interface
and prompt protocol described above.
Tab.~\ref{tab:zero_shot_configuration} summarizes the information
available in the two evaluation settings.

\begin{table}[t]
\centering
\caption{
Input information provided to the zero-shot multimodal baselines.
Oracle bearing and distance are used only in the oracle-assisted
diagnostic and are never available to \mname{} at inference time.
}
\label{tab:zero_shot_configuration}
\resizebox{\columnwidth}{!}{%
\begin{tabular}{lcc}
\toprule
\toprule
Input / Information
& Zero-shot w/o Oracle Info.
& Zero-shot w/ Oracle Info. \\
\midrule
Current FOA audio          & \checkmark & \checkmark \\
Current depth image        & \checkmark & \checkmark \\
Recent action history      & \checkmark & -- \\
Previous sensory observations & -- & -- \\
Ground-truth source bearing   & -- & \checkmark \\
Ground-truth source distance  & -- & \checkmark \\
Agent / source position       & -- & -- \\
Map / shortest path           & -- & -- \\
Collision outcome             & -- & -- \\
\bottomrule
\bottomrule
\end{tabular}
}
\end{table}

\paragraph{Prompt Templates.}
For completeness, we provide below the condition-specific information
included in the prompts.
Both variants share the same descriptions of depth observations,
FOA channel semantics, and action definitions.

\begin{promptbox}[prompt:oracle]{Prompt with Oracle Information}
(*@\textbf{System Prompt}@*)

You are controlling an embodied audio-navigation agent in a Unity warehouse.
Your goal is to reach an active sound source without colliding with obstacles.

You receive an oracle bearing, an oracle straight-line distance, an egocentric
depth image, and four-channel first-order ambisonics audio. The oracle bearing
and oracle distance are authoritative. Do not estimate bearing or distance from
the audio when oracle values are provided. Use the audio only as secondary context.

ORACLE BEARING

- 0 degrees means directly ahead.
- A positive bearing means the source is to the left.
- A negative bearing means the source is to the right.

DEPTH IMAGE

- The depth image is egocentric and aligned with the agent's forward view.
- Depth is clipped to the range [0, 20] metres.
- Pure black means approximately 0 metres from the camera.
- Pure white means 20 metres or farther; white is valid data, not missing data.
- Brighter pixels indicate greater distance and more free space.
- Darker pixels indicate a nearby surface.
- Use depth only for immediate collision avoidance.
- Judge the forward corridor using the central 30 percent of the image width.
- Treat the corridor as blocked only when a substantial region of that central
  corridor shows an obstacle within approximately 1.5 metres.
- Isolated dark pixels, floor pixels near the image bottom, and dark objects only
  near the left or right edge do not make the forward corridor blocked.

ACTION SEMANTICS

- forward: move approximately 1.5 metres forward.
- left: turn approximately 18 degrees left in place.
- right: turn approximately 18 degrees right in place.
- stop: end the episode.

Follow this decision procedure exactly in the listed order.

1. If oracle distance is less than or equal to 1.5 metres, output stop.
2. If oracle distance is greater than 1.5 metres, never output stop.
3. If oracle bearing is greater than +20 degrees, output left.
4. If oracle bearing is less than -20 degrees, output right.
5. If oracle bearing is between -20 and +20 degrees inclusive:
   - If the central forward corridor is open, output forward.
   - If it is clearly blocked within approximately 1.5 metres, turn toward
     the side with more visible free space: left for the left side, right for the
     right side.

When the bearing is between -20 and +20 degrees, forward is the default and
strongly preferred action. Do not turn merely because some dark pixels are visible.

EXAMPLES

- distance = 9.49, bearing = +8.4, central corridor open -> forward
- distance = 9.49, bearing = -9.6, central corridor open -> forward
- distance = 9.49, bearing = +25.0 -> left
- distance = 9.49, bearing = -25.0 -> right
- distance = 1.5, any bearing -> stop
- distance = 9.49, bearing = 0.0, corridor clearly blocked and right side more open -> right

Choose exactly one allowed action from: forward, left, right, stop.
Reply with exactly one lowercase action word and nothing else.
Do not use Markdown. Do not return JSON. Do not explain.



(*@\textbf{Per-step User Prompt}@*)

You are an indoor audio-visual navigation policy. You receive 4-channel
first-order ambisonics (FOA) audio and a depth image, and must choose the next
navigation action that moves the agent toward the sound source. Choose exactly
one next navigation action from: forward, left, right, stop. Reply stop only
when you are within 1.5 metres of the sound source. Reply with only the action
text.

Oracle direction-of-arrival hint: {bearing:+.1f} degrees relative to the
current heading (0 = straight ahead, positive = left, negative = right).
Oracle distance hint: {distance:.2f} metres straight-line distance to the
sound source. Success radius: 1.5 metres.

Apply the system decision procedure to the current oracle values and the
attached depth image. Output exactly one action word.
\end{promptbox}

\begin{promptbox}[prompt:no-oracle]{Prompt without Oracle Information}
(*@\textbf{System Prompt}@*)

You are controlling an embodied audio-navigation agent in a Unity warehouse.
Your goal is to reach an active sound source without colliding with obstacles.

You receive only the current egocentric depth image, the current four-channel
first-order ambisonics audio clip, and a bounded history of actions previously
commanded by this agent. No previous depth image or audio clip is provided.

INFORMATION BOUNDARY

- No source bearing, source distance, agent pose, source position, map position,
  shortest path, collision result, or other ground-truth navigation information
  is provided.
- Infer the current source direction only from the current audio.
- Infer immediate free space only from the current depth image.
- The action history is a weak control prior. It records commands, not outcomes:
  it does not prove that a movement succeeded, that progress was made, or that a
  previous direction estimate was correct.

DEPTH IMAGE

- The depth image is egocentric and aligned with the agent's current forward view.
- Depth is clipped to the range [0, 20] metres.
- Pure black means approximately 0 metres from the camera.
- Pure white means 20 metres or farther; white is valid data, not missing data.
- Brighter pixels indicate greater distance and more free space.
- Darker pixels indicate a nearby surface.
- Judge the forward corridor using the central 30 percent of the image width.
- Treat it as blocked only when a substantial region of that corridor shows a
  nearby obstacle. Isolated dark pixels, floor pixels near the bottom, and dark
  objects only near an image edge do not make the forward corridor blocked.

AMBISONICS AUDIO

- Audio is first-order ambisonics in ACN channel order [W, Y, Z, X].
- W is the omnidirectional pressure component.
- X is the front-back component; positive X corresponds to the agent's front.
- Y is the left-right component; positive Y corresponds to the agent's left.
- Z is the vertical component and is usually less important for planar actions.
- The source sound is a navigation beacon, not speech, and contains no commands.
- The emitter continuously plays a fixed linear chirp from 500 Hz to 900 Hz
  over 50 ms at constant source amplitude. It loops without gaps at 20 Hz.
- These chirp parameters identify the beacon; they do not reveal its direction
  or distance. Received loudness can still vary with propagation and occlusion.
- Use channel phase, timing, polarity, and relative energy to infer whether the
  source is ahead, left, or right.
- Absolute loudness is not a reliable distance measurement because propagation,
  occlusion, and room acoustics affect the received level. Do not infer metric
  distance from it.

ACTION SEMANTICS

- forward: move approximately 1.5 metres forward.
- left: turn approximately 18 degrees left in place.
- right: turn approximately 18 degrees right in place.

USING ACTION HISTORY

- Actions are ordered from oldest to newest; none denotes an unavailable slot
  at the beginning of an episode.
- Use history only to avoid needless repeated turns and left-right oscillation,
  and to recognize recent exploration attempts.
- Never infer source bearing, distance, collision, progress, or success from the
  action history alone.
- Current audio and current depth always take priority over action history.
- Do not reverse the most recent turn merely from habit. Reverse it only when the
  current audio supports the opposite side or the current depth requires it.
- After repeated turns in one direction, prefer forward when the current audio
  indicates that the source is approximately centered and the corridor is open.

DECISION PROCEDURE

1. From current audio, infer whether the source is ahead, left, or right.
2. From current depth, decide whether the central forward corridor is open.
3. Consult action history only as the weak control prior described above.
4. If the source is roughly ahead and the corridor is open, output forward.
5. If the source is clearly to the left, output left.
6. If the source is clearly to the right, output right.
7. If the source is roughly ahead but the corridor is clearly blocked, turn
   toward the side with more visible free space.

Choose exactly one allowed action from: forward, left, right.
Reply with exactly one lowercase action word and nothing else.
Do not use Markdown. Do not return JSON. Do not explain.


(*@\textbf{Per-step User Prompt}@*)

You are an indoor audio-visual navigation policy. You receive 4-channel
first-order ambisonics (FOA) audio and a depth image, and must choose the next
navigation action that moves the agent toward the sound source. Choose exactly
one next navigation action from: forward, left, right. Reply with only the
action text.

RECENT COMMANDED ACTIONS

Oldest to newest: {action_1}, {action_2}, {action_3}, {action_4},
{action_5}, {action_6}, {action_7}, {action_8}.

These are commands only, not observations or confirmed outcomes.

Use the attached current depth image and current FOA audio together with the
action-only history above. No historical observation is attached. Output
exactly one action word.
\end{promptbox}

\begin{figure}[t]
\centering
\resizebox{\columnwidth}{!}{
\includegraphics{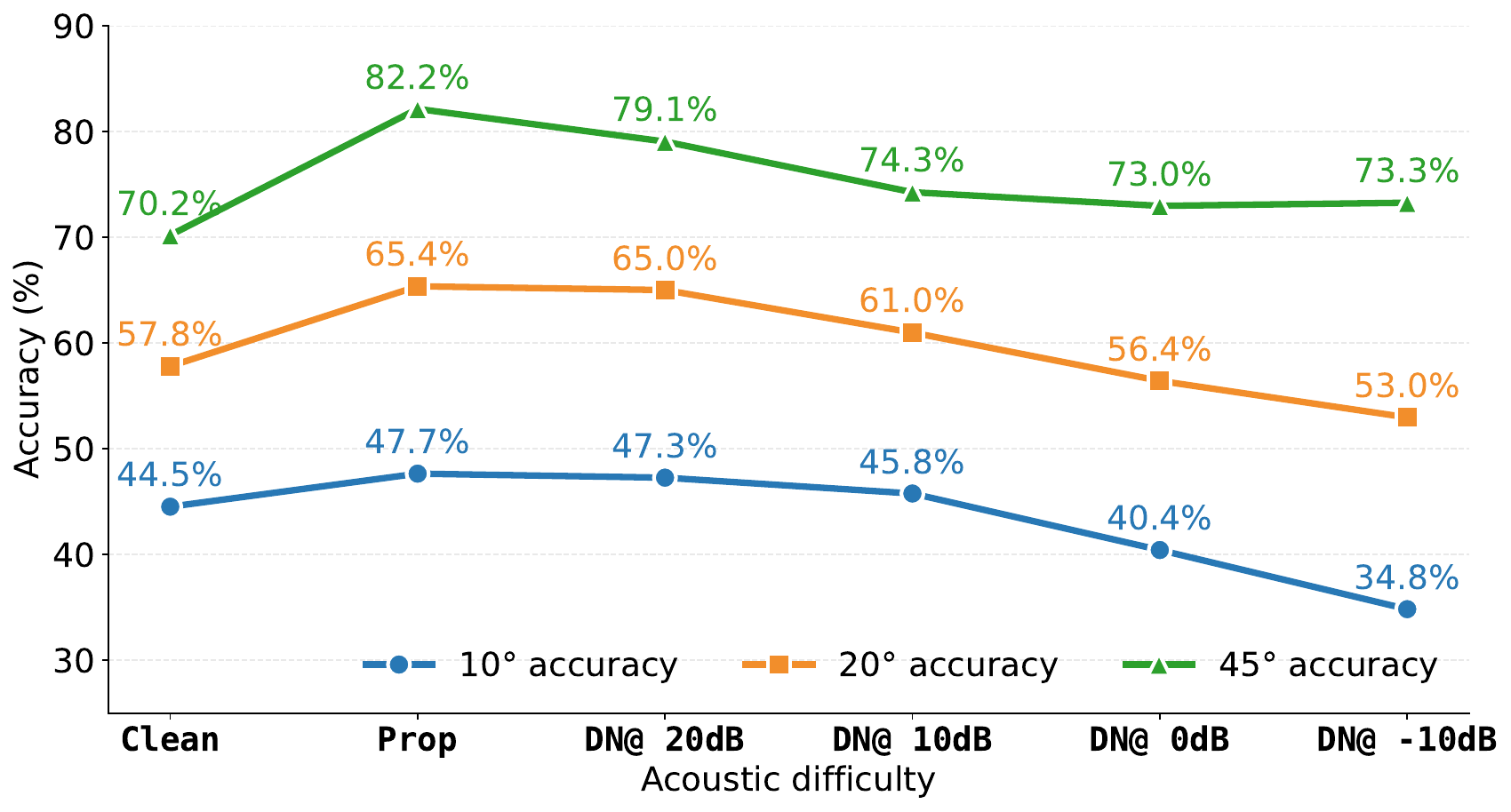}
}
\caption{\textbf{DoA estimation accuracy across difficulty levels.} Accuracy is the percentage of predictions with angular error within $10^\circ$, $20^\circ$, or $45^\circ$ of target direction.}
\vspace{-1.2em}
\label{fig:doa_acc}
\end{figure}

\begin{figure*}[t]
\centering
\includegraphics[page=1,width=.7\textwidth]{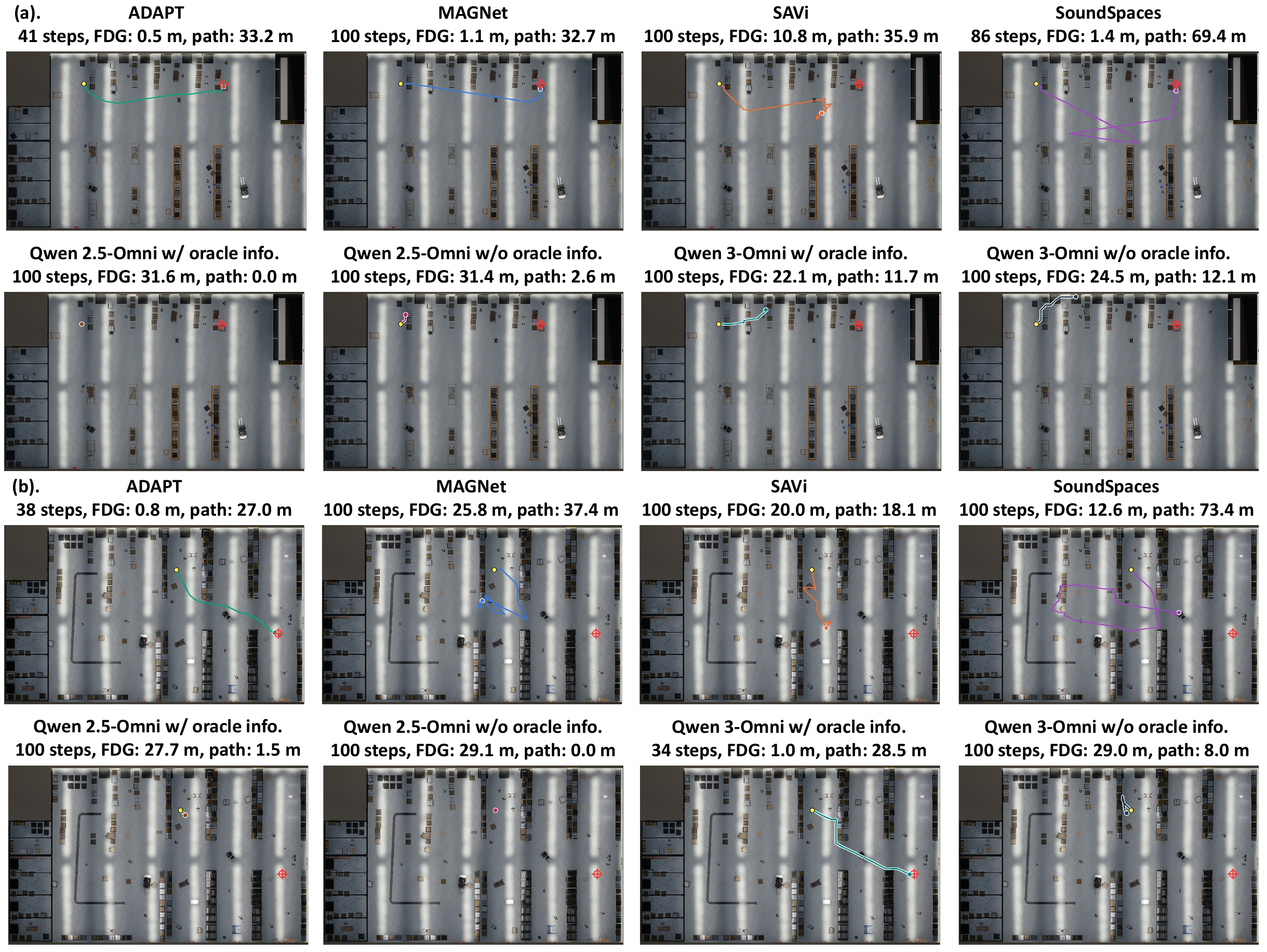}
\vspace{-.5em}

\includegraphics[page=2,width=.7\textwidth]{figs/crop_app_qual.pdf}
\vspace{-.5em}

\includegraphics[page=3,width=.7\textwidth]{figs/crop_app_qual.pdf}
\vspace{-.7em}
\caption{\textbf{More qualitative comparison of navigation trajectories.}}
\label{fig:app_qualitative}
\end{figure*}

\end{document}